\ifdefined\pdfminorversion
\fi
\documentclass[twocolumn,10pt]{wlscirep}
\usepackage[T1]{fontenc}
\usepackage{array}
\usepackage{tabularx}
\usepackage{colortbl}
\usepackage{mathtools}
\usepackage{amsthm}
\usepackage{algorithm}
\usepackage{algorithmic}
\usepackage{xspace}
\usepackage{amsmath, amsfonts}
\graphicspath{{figures/}}

\titlespacing*{\subsection}{0pc}{2.5ex plus 3pt minus 4pt}{0pt}%

\newtheoremstyle{npjdef}{3pt}{3pt}{\normalfont}{}{\bfseries}{.}{0.5em}{}
\theoremstyle{npjdef}
\newtheorem{definition}{Definition}
\newtheorem{proposition}{Proposition}
\newtheorem{remark}{Remark}

\definecolor{mygray}{gray}{.90}
\definecolor{myhead}{rgb}{0.86,0.90,0.96}
\definecolor{oursrow}{rgb}{0.93,0.92,0.985}% light violet, the "ours" family of the figures
\newcommand{\oursrow}{\rowcolor{oursrow}}
\newcolumntype{Y}[1]{>{\hsize=#1\hsize\centering\arraybackslash}X}
\newcommand{\up}{{\footnotesize$\uparrow$}}
\newcommand{\down}{{\footnotesize$\downarrow$}}
\newcommand{\headrow}{\rowcolor{myhead}}
\newcommand{\best}[1]{\textbf{#1}}

\newcommand{\ours}{Wormhole}
\newcommand{\lold}{l_{\rm old}}
\newcommand{\lnew}{l_{\rm new}}
\newcommand{\cold}{c_{\rm old}}
\newcommand{\cnew}{c_{\rm new}}
\newcommand{\pold}{p_{\rm old}}
\newcommand{\pnew}{p_{\rm new}}
\newcommand{\xold}{x^{\rm old}}

\newcommand{\dOT}{d^{\rm OT}}
\newcommand{\Pesc}{P_{\rm esc}}
\newcommand{\Tesc}{T_{\rm esc}}
\newcommand{\E}{\mathbb{E}}
\newcommand{\Ccal}{\mathcal{C}}
\newcommand{\Bcal}{\mathcal{B}}
\newcommand{\Dcal}{\mathcal{D}}

\title{Escaping the instruction black hole of vision-language-action policies via optimal-transport wormholes}

\author[1,2,3]{Xuetao Li}
\author[3]{Chen Liu}
\author[4]{Jing Wu}
\author[5]{Sheng Liu}
\author[1,2,*]{Miao Li}

\affil[1]{School of Computer Science, Wuhan University, Wuhan, China}
\affil[2]{School of Robotics, Wuhan University, Wuhan, China}
\affil[3]{Style3D Research, China}
\affil[4]{State Key Laboratory of Intelligent Vehicle Safety Technology, Changan Automobile, China}
\affil[5]{Institute of Technological Sciences, Wuhan University, Wuhan, China}
\affil[*]{Correspondence: Miao Li (miao.li@whu.edu.cn)}

\begin{abstract}
Vision-language-action (VLA) policies are trained to follow a fixed instruction, yet people change their minds while the robot acts. We identify the instruction black hole: once a VLA policy commits to a task, new instructions are absorbed and the robot completes the old task or freezes. We quantify it with a transmission ratio and an event horizon and prove it is a blind spot of static imitation, which cannot distinguish a policy that listens from a deaf one once the execution state reveals the task. We propose Wormhole, a training-free plug-in that relocates the robot instead of re-instructing it: optimal transport selects an approach-phase state of the new task, a rate-limited traversal carries the robot there, and the unmodified policy takes over. On LIBERO, Wormhole raises committed-switch success of $\pi_0$ and $\pi_{0.5}$ from at most 1.2\% to 93.1\%, matching or exceeding scripted retreat-and-restart with one fifth of its action discontinuity.
\end{abstract}

\begin{document}

\flushbottom
\maketitle
\thispagestyle{empty}

\section*{Introduction}

Vision-language-action (VLA) policies have turned language into a universal interface for robot control. By mapping camera images and an instruction directly to short chunks of actions, models such as RT-2, OpenVLA, $\pi_0$, and $\pi_{0.5}$ execute a wide range of manipulation tasks without task-specific controllers~\cite{rt2,openvla,pi0,pi05}. Their training and evaluation, however, rest on a quiet assumption: the instruction is fixed for the whole episode. People do not work that way: a user who asked for the apple may, a moment later, want the banana, and a robot that shares space with people must be able to change its mind when they change theirs.

We began with a simple probe. The publicly released $\pi_{0.5}$ policy fine-tuned on LIBERO~\cite{libero,openpi} was asked to put the alphabet soup in the basket; while it was reaching, we told it to fetch the salad dressing instead. Issued at control step 8, the new instruction was followed in 7 of 10 trials. Issued at step 16, with both objects still in plain view, it was followed only once. In 7 of the 10 trials, the robot put the alphabet soup in the basket as if nothing had been said. The new instruction caused no hesitation and no error; it simply vanished. We call this phenomenon the \textbf{instruction black hole}. Like its astrophysical namesake, it has an \emph{event horizon}, a point in the execution beyond which a new instruction can no longer escape the pull of the ongoing task.

\begin{figure*}[!tp]
\centering
\includegraphics[width=\textwidth]{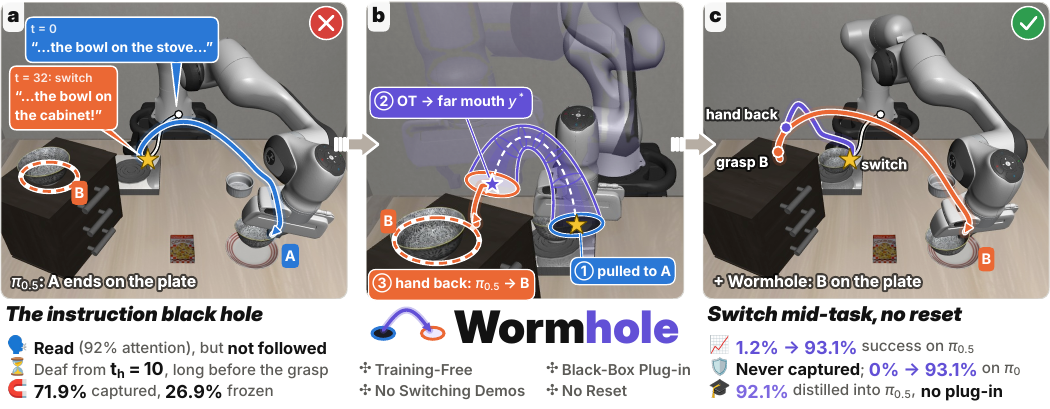}
\caption{\textbf{The instruction black hole and the \ours{} escape.} (a) $\pi_{0.5}$ on LIBERO is told to fetch the bowl on the stove (A); at $t_s=32$ (star), with the arm at A, it is told to fetch the identical bowl on the wooden cabinet (B). The policy keeps the old task and carries A to the plate (frame $t=109$ and the end-effector path; white: before the switch). The instruction is read but not used: at the last query before the switch the action queries still give it 92\% of the attention they gave it at $t=0$ (Fig.~\ref{fig:overview}). Beyond the horizon $t_h=10$, 30--40 steps before the grasp (Fig.~\ref{fig:escape}(a)), vanilla $\pi_{0.5}$ is captured in 71.9\% and freezes in 26.9\% of the committed switches. (b) \ours{} at the switch, drawn in the scene: (1) the policy's continuations of the old task lead to A; (2) optimal transport between these exits and approach-phase states of successful executions of B selects the far mouth $y^*$ (violet star), and a rate-limited traversal carries the arm there (tube: schematic; translucent: the arm when it is handed back); (3) the unmodified policy, queried under the new instruction, grasps B (orange). (c) The same switch with \ours{}, from the same state and noise seed as (a): B ends on the plate, without resetting the arm. Committed-switch success on the test split rises from 1.2\% to 93.1\% on $\pi_{0.5}$ and from 0\% to 93.1\% on $\pi_0$, with no capture (Table~\ref{tab:main}; retreat-and-restart, which resets the arm, reaches 90.0\% and 83.8\%), and reaches 92.1\% for a $\pi_{0.5}$ trained on the switching data of a relocation teacher, with no plug-in at test time (see ``Distilling the wormhole into the policy'').}
\label{fig:teaser}
\end{figure*}

The black hole is not a quirk of one checkpoint or one scene. On two LIBERO tasks that share a layout and differ only in which of two identical bowls to fetch, $\pi_0$ and $\pi_{0.5}$ follow a switch issued at the start of the episode in 80 of 80 trials, but only 2 of 80 switches issued 15 steps later, 20--40 steps before the gripper closes on the bowl. On LIBERO-Goal, $\pi_{0.5}$ follows every switch issued at the start, but only 12 of 48 switches to a different object issued 20 steps later and 1 of 48 issued after 60 steps (see Results); independent measurements report the same collapse for $\pi_0$ and OpenVLA-OFT~\cite{switchvla}. The phenomenon has three signatures: \emph{capture}, in which the old task runs to completion (Fig.~\ref{fig:teaser}(a)); \emph{freezing}, in which the robot stalls near the horizon; and \emph{tidal disruption}, in which an escape does occur but tears the action stream apart with jerky motion. Our first research question is thus: \textsl{\textbf{I)} Why do VLA policies stop listening once they have committed to a task?}

To address \textsl{\textbf{I)}}, we show that the black hole reflects a lack of supervision, not of capacity. In static demonstrations, each trajectory is generated under a single instruction. Once the robot moves toward its target, the images and the arm's own motion reveal which task is being performed, and the instruction becomes redundant. We prove that in this \emph{committed region} the imitation objective is exactly invariant to whether the policy reads the instruction (Proposition~\ref{prop:blind}): a policy that listens and a policy that is deaf attain the same loss, and the documented preference of deep networks for the simplest sufficient cue~\cite{dehaan,geirhos,shah,xing} favors deafness. Hence more demonstrations of the same kind cannot close the black hole, hierarchical replanning~\cite{saycan,yay,hirobot} cannot either (the low-level policy must still listen), and amplifying the instruction at inference time can only revive a signal that is weak, not one that was never learned. This leads to our second question: \textsl{\textbf{II)} How can a policy escape the black hole without switching demonstrations?}

To address \textsl{\textbf{II)}}, we turn the cause into the cure. If the state, not the instruction, decides what the policy does inside a committed region, then a robot placed inside the committed region of the \emph{new} task will execute the new task, whatever it reads (Proposition~\ref{prop:reloc}). We propose \textbf{\ours{}}, a training-free plug-in that transduces the new instruction into such a state and carries the robot there. At the switch, \ours{} samples how the old task would continue (mouth A), collects states visited by successful executions of the new instruction before their grasp (mouth B), and couples the two with optimal transport (OT)~\cite{peyre,cuturi} under the expected time to complete the new task. The coupling defines the far mouth of the wormhole, a state inside the basin of the new task; a short, rate-limited traversal brings the robot there, and the unmodified policy finishes (Fig.~\ref{fig:teaser}(b)). \ours{} never asks the base policy to listen inside the black hole: the same commitment that caused the black hole now holds the robot on the new task.

We evaluate the black hole and \ours{} on LIBERO with $\pi_0$ and $\pi_{0.5}$, in a switching protocol in which every comparison is a paired counterfactual from the same initial state and all choices were fixed on a development split before a single test run. The horizon is abrupt, precedes physical commitment by tens of steps, and is absent where the execution state does not yet identify the task. Without training, \ours{} raises committed-switch success from 0\% to 93.1\% on $\pi_0$ and from 1.2\% to 93.1\% on $\pi_{0.5}$ (Fig.~\ref{fig:teaser}(c)) and is never captured, whereas inference-time remedies reach at most 11.2\% and an action-space bridge that steers the motion without relocating the robot at most 33.1\%. A scripted retreat to the initial pose followed by a restart succeeds in 83.8\% ($\pi_0$) and 90.0\% ($\pi_{0.5}$): \ours{} matches it on $\pi_{0.5}$ and exceeds it on $\pi_0$, with one fifth of its action discontinuity and without a reset pose. Finally, the switching data that the plug-in generates let a $\pi_{0.5}$ learn to listen by itself. Our contributions are:

\begin{itemize}
\item \textbf{Instruction black hole.} We identify the instruction black hole of VLA policies and make it measurable through the transmission ratio, the event horizon, and three behavioral signatures, together with an evaluation protocol for mid-execution switching on LIBERO.
\item \textbf{Identifiability theory.} We prove that the black hole is a blind spot of static imitation (Proposition~\ref{prop:blind}) and that the same blind spot locks a policy onto any task whose committed region it is placed in (Proposition~\ref{prop:reloc}), which turns escape into relocation.
\item \textbf{\ours{}.} A training-free plug-in that transduces the instruction from language into state-space geometry: OT between the policy's sampled continuations and states of successful new-task executions selects the far mouth, and a rate-limited traversal delivers the robot into the basin of the new task, without switching demonstrations or access to the weights.
\item \textbf{Evidence, and a self-trained counterpart.} Paired-counterfactual escape curves on three LIBERO suites and two backbones characterize the horizon and its failure modes, and mechanistic measurements tie it to task identifiability. On a test split evaluated once, \ours{} escapes 93.1\% of committed switches on both backbones, and a hand-over analysis across all methods confirms that the escape is decided by relocation. Trained on switching data from an oracle relocation teacher, a $\pi_{0.5}$ whose instruction reaches the actions only through a grounding token follows 92.1\% of committed switches without any plug-in, significantly above behavior cloning trained on the same data.
\end{itemize}

Large-scale imitation learning has produced generalist policies conditioned on language and vision~\cite{gato,bcz,mandlekar,rt1,oxe,palme,vima,octo}. VLA models fine-tune a vision-language backbone to emit actions, as discrete tokens in RT-2 and OpenVLA~\cite{rt2,openvla,oft} or through a flow-matching action expert in $\pi_0$ and $\pi_{0.5}$~\cite{pi0,pi05,instructvla}; they predict action chunks~\cite{act,diffusionpolicy,rdt,flowmatching}, and bidirectional decoding and real-time chunking~\cite{bid,rtc} make consecutive chunks consistent, which by design favors continuing the motion under way. All assume one instruction per episode. Systems that redirect robots with language---SayCan~\cite{saycan}, Interactive Language~\cite{interactivelang}, online corrections~\cite{sharma,cui}, YAY Robot and Hi Robot~\cite{yay,hirobot}, RACER~\cite{racer}, ReKep~\cite{rekep}, and completion-aware hand-overs~\cite{cab}---decide \emph{what} the robot should do next and assume that the low-level policy will follow; the black hole violates exactly this assumption. Closest to our setting, HAMLET adds history awareness~\cite{hamlet} and SwitchVLA trains a dedicated policy on contact-phase-segmented demonstrations with rollback and advance behaviors that interpolate toward a canonical start pose~\cite{switchvla}. \ours{} explains why switching fails, plugs into existing backbones, and, instead of retreating to a canonical pose, carries the robot forward to an OT-matched state in the basin of the new task.

Imitation policies exploit nuisance correlates of the expert's actions (causal confusion~\cite{dehaan}), deep networks prefer the simplest sufficient features~\cite{geirhos,shah}, generalist policies inherit such shortcuts~\cite{xing}, and compounding errors expose them at deployment~\cite{dagger}. VLA policies often disregard their instructions in static evaluation~\cite{liberopro,liberoplus}; we show that the same mechanism has a dynamic form---language becomes redundant as execution unfolds---and prove that static imitation cannot detect it. OT provides geometry-aware correspondences between distributions~\cite{villani,peyre,cuturi} and has matched agent and expert trajectories for imitation rewards~\cite{watchmatch}; flow matching, interpolants, rectified flows, minibatch OT couplings, and Schr\"odinger bridges~\cite{flowmatching,interpolants,rectified,otcfm,multisample,dsb,i2sb} learn paths between distributions, and SDEdit regenerates a sample under a new condition~\cite{sdedit}. Bridging the old action chunk into the new one is the natural way to apply these tools to switching, and we evaluate such an action-space bridge; \ours{} instead transports between where the old task is taking the robot and where the new task is executed from, and uses the plan itself as the carrier of the instruction.

\section*{Results}

\subsection*{Mid-execution switching and its measurement}
At control step $t$, a VLA policy receives observations $o_t$, proprioception $q_t$, an instruction $l$, and an execution context $s_t$ that summarizes recent motion,
\begin{equation}
\label{eq:context}
s_t=\{q_{t-r:t},\,a_{t-r:t-1},\,g_{t-r:t},\,e_{t-r:t}\},
\end{equation}
with history length $r$, gripper state $g_t$, and end-effector pose $e_t$. We write $\chi_t=(o_t,q_t,s_t)$ for the execution state and $c_t=(\chi_t,l)$ for the full condition. A flow-based policy generates an action chunk $A_t=\{a_{t+1},\ldots,a_{t+K}\}$ by integrating a learned velocity field over a generative time $\tau\in[0,1]$,
\begin{equation}
\label{eq:flow}
\begin{gathered}
\frac{{\rm d}z_\tau}{{\rm d}\tau}=v_\theta(z_\tau,\tau\mid c_t),\\
z_0\sim\mathcal{N}(0,I),\qquad A_t=D(z_1),
\end{gathered}
\end{equation}
where $D(\cdot)$ maps the latent chunk to actions; autoregressive policies decode discretized actions, and we treat their de-tokenized chunks in the same space. In mid-execution switching, an episode starts under $\lold$; at a switch step $t_s$ the user issues $\lnew\neq\lold$, and the condition changes from
\begin{equation}
\label{eq:switch}
\cold=(\chi_{t_s},\lold)\quad\text{to}\quad \cnew=(\chi_{t_s},\lnew).
\end{equation}
Both instructions are executable from the initial state. A trial succeeds only if the rollout reaches the terminal state of $\lnew$; completing $\lold$ or timing out counts as failure. The naive response, implicitly adopted by current policies, is to replace $\cold$ by $\cnew$ and continue.

Let $\pold(\cdot\mid\chi)$ be the distribution of the policy's continuations under $\lold$ and $\pnew(\cdot\mid\chi)$ the distribution of reachable entry segments of $\lnew$ from $\chi$, i.e., short segments of successful $\lnew$ executions whose start states are compatible with $\chi$ (Methods). Their means define the \emph{required displacement} $d^\star(\chi)=\E_{\pnew}[x]-\E_{\pold}[x]$, and the policy's response is $\Delta_\theta(\chi)=\E[z_1\mid\chi,\lnew]-\E[z_1\mid\chi,\lold]$.

\begin{definition}[Instruction transmission ratio, black hole, and horizon]\label{def:rho}
The transmission ratio of $\pi_\theta$ at $\chi$ for the switch $\lold\to\lnew$ is
\begin{equation}
\label{eq:rho}
\rho_\theta(\chi)=\frac{\langle\Delta_\theta(\chi),\,d^\star(\chi)\rangle}{\|d^\star(\chi)\|^2},
\end{equation}
the fraction of the required redirection that the policy expresses along the right direction ($1$: full transmission; $0$: complete absorption). For $\epsilon\in(0,1)$, the $\epsilon$-black hole is $\Bcal_\epsilon=\{\chi:\rho_\theta(\chi)\le\epsilon\}$, its boundary is the event horizon, and along a rollout under $\lold$ the horizon time is $t_h=\inf\{t:\chi_t\in\Bcal_\epsilon\}$.
\end{definition}

The behavioral counterpart of $\rho_\theta$ is the \emph{escape probability} $\Pesc(t_s)$, the probability that a switch issued at $t_s$ succeeds, estimated by sweeping $t_s$ (see ``Anatomy of the black hole''). An instruction that falls into the black hole manifests as \emph{capture} (the rollout completes $\lold$), \emph{freezing} (the robot stalls or redirects too late), or \emph{tidal disruption} (the redirection succeeds with discontinuous, jerky actions), each with its own metric (defined in Methods).

\subsection*{Why black holes form}
Let $\Dcal$ be a distribution of static demonstrations over triples $(\chi,l,A)$, each episode generated under a single instruction, and let a policy minimize
\begin{equation}
\label{eq:imitation}
J_\Dcal(\pi)=\E_{(\chi,l,A)\sim\Dcal}\big[\ell\big(\pi(\cdot\mid\chi,l),A\big)\big],
\end{equation}
which covers negative log-likelihood, token-level cross-entropy, and the conditional flow-matching loss of $\pi_0$ and $\pi_{0.5}$.

\begin{definition}[Committed region]\label{def:committed}
The committed region of $\Dcal$ is the set of execution states at which the demonstrations leave no ambiguity about the instruction, $\Ccal=\{\chi:\ p_\Dcal(l\mid\chi)=\mathbf{1}[l=g(\chi)]\ \text{for some}\ g(\chi)\}$.
\end{definition}

Early in an episode the robot passes through states shared by many tasks, so the policy must read the language to act; once it moves toward a particular object, its state reveals the task and $\chi$ enters $\Ccal$.

\begin{proposition}[Static imitation is blind to the black hole]\label{prop:blind}
Let $\Ccal$ be the committed region of $\Dcal$ and $L$ the instruction random variable. Then (i) $I_\Dcal(A;L\mid\chi)=0$ for every $\chi\in\Ccal$; (ii) for every policy $\pi$, the deafened policy $\tilde\pi(\cdot\mid\chi,l)=\pi(\cdot\mid\chi,g(\chi))$ for $\chi\in\Ccal$ and $\tilde\pi=\pi$ elsewhere satisfies $J_\Dcal(\tilde\pi)=J_\Dcal(\pi)$; and (iii) $\rho_{\tilde\pi}(\chi)=0$ for every $\chi\in\Ccal$ with $d^\star(\chi)\neq0$: every switch issued in $\Ccal$ is completely absorbed by $\tilde\pi$.
\end{proposition}

The proof is in Supplementary Note~1. The imitation objective has a flat direction exactly where switching matters: on $\Ccal$, any degree of deafness is free, and the deaf solution, which ignores a redundant input, is the simplest one available.

\begin{remark}[Scale does not help; testable predictions]\label{rem:predictions}
The invariance depends only on the support of $\Dcal$, not on its size: more demonstrations of the same kind leave $\Ccal$ unchanged, and only data that make committed states ambiguous---switching data---can remove the blind spot (``Distilling the wormhole into the policy'' tests this). The analysis further predicts that (P1) the horizon time $t_h$ coincides with the time at which the task becomes decodable from the execution state alone; (P2) inference-time interventions that amplify the instruction help near the horizon, where a residual signal survives, but fail deep inside it; and (P3) a robot delivered into the committed region of the new task executes it whether or not it reads the instruction, so success after a switch is determined by where the robot is handed back to the policy. The experiments below test (P1)--(P3). If the initial scene already identifies the task, $\Ccal$ covers almost the whole episode and the policy may ignore language from the start: the static language neglect of refs.~\citen{liberopro,liberoplus} is the limiting case of the black hole.
\end{remark}

\subsection*{Overview of \ours{}}
\begin{figure*}[!tp]
\centering
\begingroup
\setlength{\unitlength}{\dimexpr\textwidth*1000/639224\relax}%
\definecolor{figdark}{rgb}{0.122,0.137,0.161}%
\begin{picture}(639.224,523.126)
\put(0,0){\includegraphics[width=\textwidth]{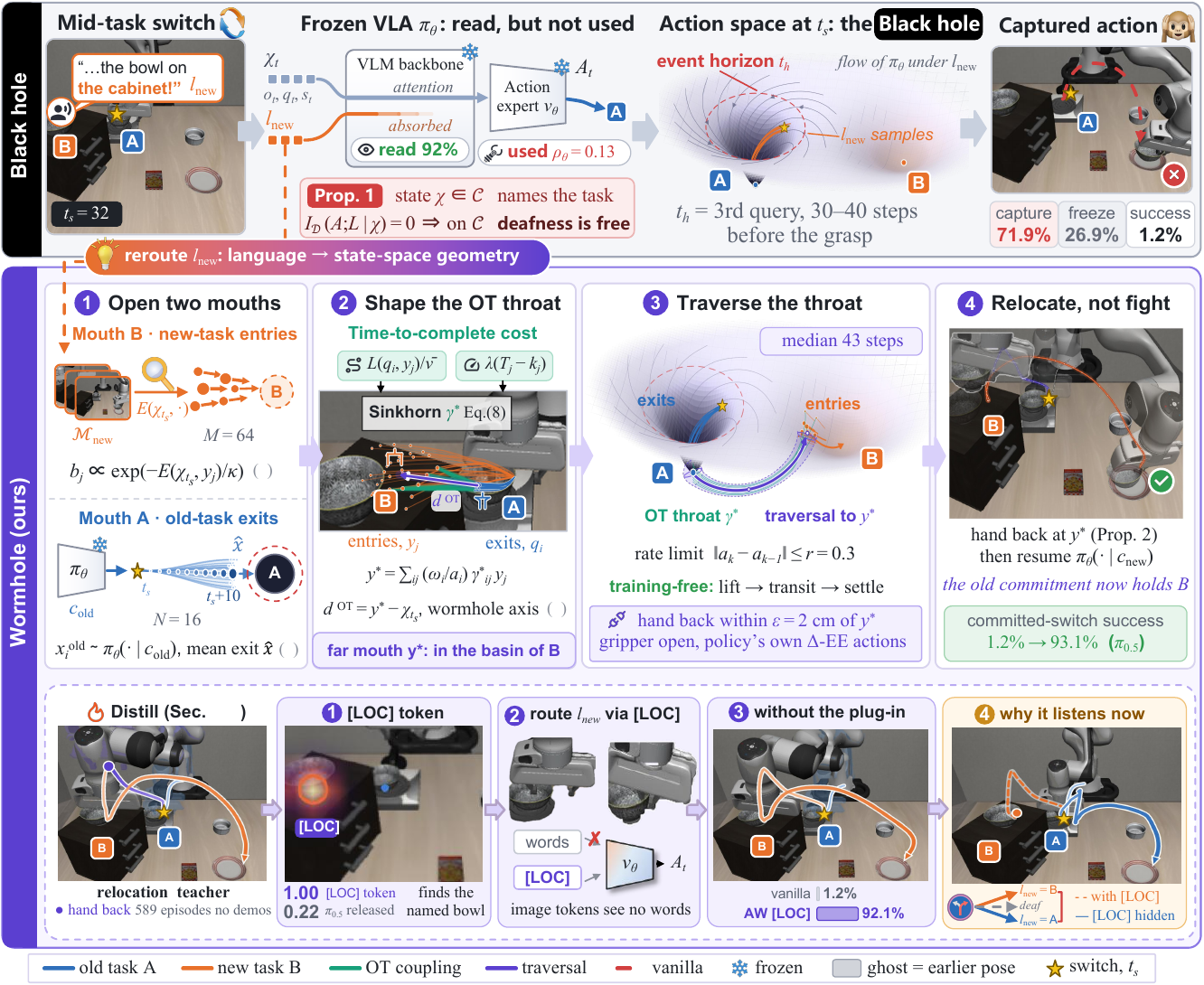}}
% equation numbers in the empty brackets: b_j (Mouth B), mean exit (Mouth A), wormhole axis
\put(139.626,270.316){\makebox[0pt][c]{\fontsize{7.05}{8}\selectfont\rmfamily\ref{eq:pnew}}}
\put(153.426,175.272){\makebox[0pt][c]{\fontsize{7.05}{8}\selectfont\rmfamily\ref{eq:pold}}}
\put(295.735,196.832){\makebox[0pt][c]{\fontsize{7.05}{8}\selectfont\rmfamily\ref{eq:axis}}}
% "Distill (Sec.  )": sections are unnumbered here, so the bracket names the section and links to the subsection
\put(86.621,139.12){\textcolor{white}{\rule{44.814\unitlength}{10.2\unitlength}}}
\put(87.021,141.72){\fontsize{7.32}{8}\selectfont\sffamily\bfseries\textcolor{figdark}{(}\hyperref[sec:distill]{Results}\textcolor{figdark}{)}}
% clickable areas over the references the figure already prints: Eq.(8), Prop. 1, Prop. 2
\put(246.208,298.48){\hyperref[eq:ot]{\rule{0pt}{9.5\unitlength}\hspace*{22.514\unitlength}}}
\put(163.498,413.082){\hyperref[prop:blind]{\rule{0pt}{12\unitlength}\hspace*{39.004\unitlength}}}
\put(579.159,233.184){\hyperref[prop:reloc]{\rule{0pt}{10\unitlength}\hspace*{35.563\unitlength}}}
\end{picture}%
\endgroup
\caption{\textbf{The black hole and the pipeline of \ours{}} ($\pi_{0.5}$, LIBERO-Spatial pair, switch from the bowl on the stove (A) to the bowl on the cabinet (B) at $t_s=32$). \emph{Top, the black hole.} At the last query before the switch ($t=30$), the action queries still give the instruction 92\% of the attention they gave it at $t=0$, but the action expresses only $\rho_\theta=0.13$ of the required redirection: the instruction is read but not used, because in the committed region the state already names the task (Proposition~\ref{prop:blind}). Samples under $\lnew$ drain into A; the horizon $t_h$ is the third query, 30--40 steps before the grasp. Over committed switches, the vanilla policy is captured in 71.9\%, freezes in 26.9\%, succeeds in 1.2\%, and inference-time fixes reach at most 11.2\%. \emph{Middle, \ours{}.} (1) Mouth~A samples $N=16$ continuations of the old task (Eq.~\eqref{eq:pold}; dots: the mean exit $\hat x$ from $t_s$ to $t_s+10$); Mouth~B takes approach-phase states of successful executions of $\lnew$, weighted by the expected steps to complete $\lnew$ through them (Eq.~\eqref{eq:pnew}). (2) Entropic OT couples the end points of the exits (blue) with the entries (orange) and defines the far mouth $y^*$ and the axis $\dOT$ (violet; Eqs.~\eqref{eq:ot}, \eqref{eq:axis}); the photo illustrates the throat at $t=30$ with the 30 largest couplings (green). (3) A rate-limited traversal opens the gripper, lifts, transits, and settles on $y^*$ (median 43 steps). (4) From $y^*$ the unmodified policy resumes under $\lnew$ and completes B (Proposition~\ref{prop:reloc}); committed success rises from 1.2\% to 93.1\% (Table~\ref{tab:main}). \emph{Bottom, distillation (see ``Distilling the wormhole into the policy''); the robot photos are exact replays of recorded A$\to$B switches.} An oracle relocation teacher carries the arm toward B and hands it back (violet dot) to the released policy (589 episodes, no human demonstrations); on these data a $\pi_{0.5}$ with a grounding token [LOC] is trained: (1) [LOC] looks at the named bowl (pointing rate 1.00; released policy 0.22); (2) the instruction reaches the actions only through [LOC]; (3) without any plug-in, the policy follows 92.1\% of the committed test switches (behavior cloning 85.2\%); (4) on switching data deafness costs loss (Remark~\ref{rem:predictions}), and hiding [LOC] makes the policy deaf again: on the development split its committed success falls from 91--96\% to 5--20\%; in the photo it carries A to the plate (dashed: the same state with [LOC]).}
\label{fig:overview}
\end{figure*}

Inside the committed region the base policy cannot be redirected through its instruction input, so \ours{} does not ask it to listen; it changes \emph{where} the policy acts. The commitment that absorbs $\lnew$ inside the black hole of the old task would hold the robot on the new task if the robot were inside the black hole of the new one. As illustrated in Fig.~\ref{fig:overview}, at the switch \ours{} (i) opens two \emph{mouths}: the exits through which the old black hole pulls the robot, sampled from the base policy, and the entries of the new black hole, states visited by successful executions of $\lnew$; (ii) shapes the \emph{throat} by an OT problem that couples exits with entries under the expected time to complete the new task; and (iii) \emph{traverses} the throat, carrying the robot through free space to the transport-weighted entry with bounded tidal stress, after which the unmodified policy resumes under $\lnew$. \ours{} requires no training, no switching demonstrations, and no access to the weights; the procedure is listed as Algorithm~\ref{alg:escape} and detailed in Methods.

We ask whether the black hole behaves as predicted, whether \ours{} escapes it and training-free remedies can do the same, and whether a policy can learn the escape from the plug-in's own data; the protocol, switch pairs, backbones, and metrics are described in Methods. Counts are given with their trial numbers; intervals are 95\% Wilson intervals.

\subsection*{Anatomy of the black hole}
\begin{figure*}[!tp]
\centering
\includegraphics[width=\textwidth]{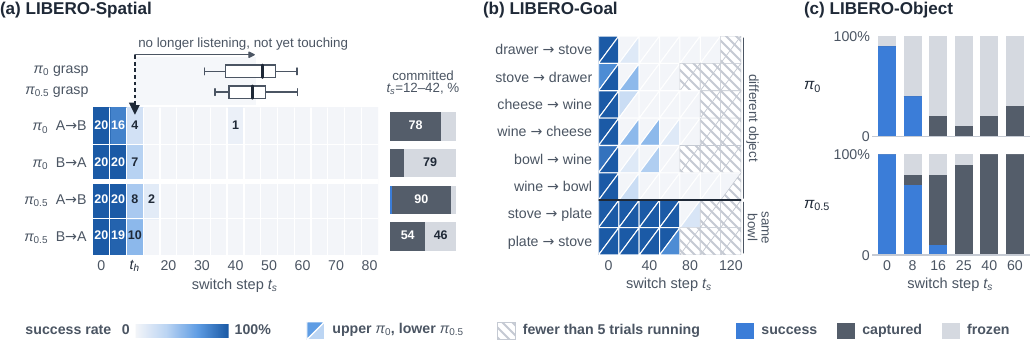}
\caption{\textbf{Behavioral anatomy of the instruction black hole.} (a) LIBERO-Spatial: successes after a switch for every switch step and condition (20 trials per cell). Above, the step at which the gripper closes in no-switch rollouts (box: interquartile range; whiskers: 5th--95th percentiles): both policies stop following a new instruction at the horizon $t_h=10$ (dashed line), tens of steps before they touch the bowl. Right, how committed switches fail ($t_s=12$--42, 80 trials per row; captured and frozen in \%). (b) LIBERO-Goal: success per ordered pair and switch step, $\pi_0$ in the upper and $\pi_{0.5}$ in the lower half of each cell (8 trials per cell; hatched: fewer than five trials still running). Switches to a different object (top six rows) collapse after commitment; switches that change only the destination (bottom two rows) are followed until the object is carried. (c) LIBERO-Object, alphabet soup$\to$salad dressing: outcome composition per switch step for $\pi_0$ (top) and $\pi_{0.5}$ (bottom), 10 trials per bar.}
\label{fig:escape}
\end{figure*}

\noindent\textbf{The horizon arrives early and abruptly.} We sweep $t_s$ from 0 to 80 in steps of five on the spatial pair (Fig.~\ref{fig:escape}(a)). Without a switch, $\pi_0$ completes both instructions in 115 of 120 rollouts. With a switch, the collapse is sharp: in the A$\to$B direction, $\pi_0$ succeeds in 20, 16, 4, and 0 of 20 trials at $t_s=0$, 5, 10, and 15 (B$\to$A: 20, 20, 7, 0), and from $t_s=15$ on a single success occurs in 560 trials. Defining the behavioral horizon $t_h$ as the first switch step at which $\Pesc$ falls to half of its value at $t_s=0$, both directions give $t_h=10$, the third query of the episode. $\pi_{0.5}$ (120 of 120 no-switch rollouts) has the same horizon: 20, 20, 8, 2 of 20 A$\to$B and 20, 19, 10, 0 B$\to$A switches at the same steps, and none of 520 switches issued from $t_s=20$ on. The horizon lies far ahead of physical commitment: the gripper closes on the bowl at a median of 37--51 steps (Fig.~\ref{fig:escape}(a), top), so the instruction stops reaching the robot 30--40 steps before the robot touches anything. Inside the horizon, the directions fail differently: for $\pi_0$, A$\to$B switches end in capture (77.5\% of committed trials) and B$\to$A switches mostly freeze (78.8\%); $\pi_{0.5}$ is captured more often (90.0\% and 53.8\%). An independent run on a second machine, before any result reported here, gave the same horizon and failure modes (Methods).

\noindent\textbf{Not specific to one pair or backbone.} On LIBERO-Goal, $\pi_{0.5}$ follows every switch issued at $t_s=0$ (64 of 64), but switches to a different object succeed in 12/48 at $t_s=20$, 7/48 at $t_s=40$, and 1/48 at $t_s=60$ (Fig.~\ref{fig:escape}(b)), whereas switches that keep the object and change its destination succeed in 48 of 48 trials up to $t_s=40$, in 14 of 16 at $t_s=60$, and collapse only at $t_s=80$, when the arm already carries the bowl toward the old destination; $\pi_0$ shows the same contrast (45/48 at $t_s=0$ but 2/48 at $t_s=20$ for a different object; 16/16 at every $t_s\le60$ for a new destination). This is what Proposition~\ref{prop:blind} predicts: the state identifies \emph{which object} as soon as the arm approaches it, but \emph{which destination} only once the arm moves toward one. On LIBERO-Object, a switch from the alphabet soup to the salad dressing succeeds in 10/10, 7/10, and 1/10 trials at $t_s=0$, 8, and 16, when 7 of 10 robots put the alphabet soup in the basket instead (Fig.~\ref{fig:escape}(c)); $\pi_0$ falls from 9/10 to 4/10 and 0/10 but mostly freezes. The reverse switch fails even at $t_s=0$ (1/10): in that scene the policy never follows the instruction to fetch the alphabet soup, the static language neglect of refs.~\citen{liberopro,liberoplus}.

\begin{figure*}[!tp]
\centering
\includegraphics[width=\textwidth]{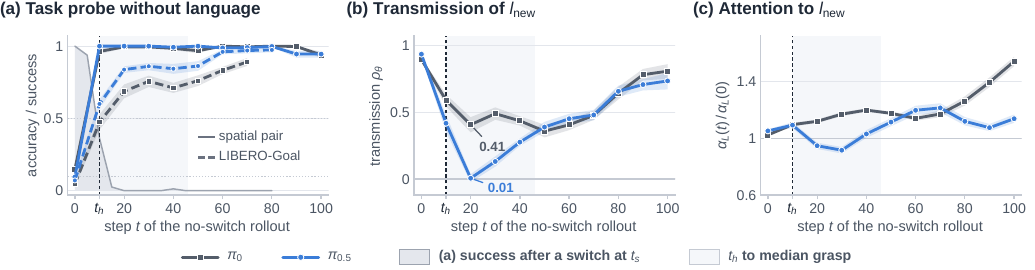}
\caption{\textbf{Mechanism of the black hole along no-switch rollouts.} (a) Accuracy of a linear task probe on instruction-free backbone features, on the spatial pair (solid) and on LIBERO-Goal (dashed); dotted lines: majority rate 0.5 and chance 0.1; gray area: success after a switch at $t_s$ on the spatial pair (both backbones and directions, 80 trials per step; Fig.~\ref{fig:escape}(a)). (b) Transmission ratio $\rho_\theta$ of the new instruction on the spatial pair. (c) Attention mass of the action queries on the instruction tokens, relative to $t=0$. The vertical line marks the horizon $t_h=10$; the shaded band extends to the median grasp step; bands around the lines are bootstrap 95\% intervals. At the horizon the task becomes visible without language and transmission collapses, while attention does not: the instruction is still read but no longer acted upon.}
\label{fig:mechanism}
\end{figure*}

\begin{figure*}[!tp]
\centering
\includegraphics[width=\textwidth]{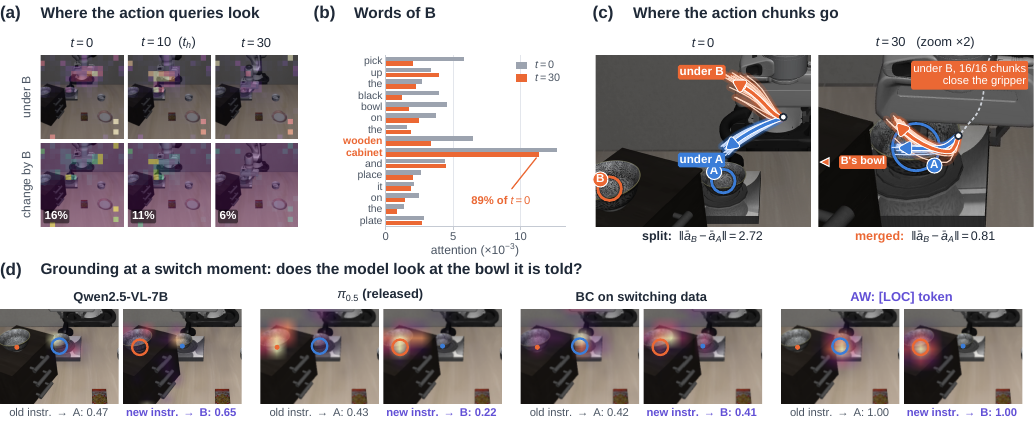}
\caption{\textbf{Inside the policy.} (a)--(c) $\pi_{0.5}$ along the no-switch rollout of A from initial state 0, from which the switch of Figs.~\ref{fig:teaser} and~\ref{fig:qualitative} is issued; at every query step the recorded observation is re-run under A and under B with four shared noise samples. (a) Attention of the action queries over the third-person image under B (top; mean over layers, heads, integration steps, and noise samples) and its change with respect to A, $|\alpha(B)-\alpha(A)|$ (bottom); the label gives the summed change as a share of the attention mass on this image. (b) Attention on the words of B at $t=0$ and $t=30$. (c) The motion commanded by 16 chunks sampled under A (blue) and under B (orange) at $t=0$ and $t=30$, drawn from the end effector into the third-person camera image and three times longer (bold arrow: mean chunk; rings: the two bowls; the view at $t=30$ is magnified twice); $\|\bar a_B-\bar a_A\|$ is the distance between the mean chunks in action units. At $t=0$ the chunks under B lift the gripper toward the cabinet while those under A descend toward the stove bowl; at $t=30$ those under B follow those under A toward the stove bowl, and all 16 close the gripper. (d) Grounding at a switch moment of the development split (A$\to$B, $t_s=12$, initial state 40): attention of Qwen2.5-VL-7B at the position that writes the box coordinates, of the released $\pi_{0.5}$ and of the BC policy at the instruction position (three best heads each), and of the [LOC] token of AW (see ``Distilling the wormhole into the policy''), under the old and the new instruction; ring: the bowl that the instruction names; number: rate at which the most attended patch lies within 1.5 patches of the named bowl over 100 switch moments (chance 0.03).}
\label{fig:inside}
\end{figure*}

\noindent\textbf{Absorption and identifiability.} Along no-switch rollouts under $\lold$ (77 for $\pi_0$, 80 for $\pi_{0.5}$), we compute the transmission ratio $\rho_\theta(\chi_t)$ of Definition~\ref{def:rho}, estimating the conditional means with four shared noise samples per instruction and $d^\star$ from state-compatible segments of no-switch $\lnew$ rollouts, together with the language attention mass $\alpha_L(t)$ of the action queries on the instruction tokens, normalized by its value at $t=0$ (Fig.~\ref{fig:mechanism}). At the start the instruction is fully transmitted ($\rho_\theta=0.89$ for $\pi_0$, $0.93$ for $\pi_{0.5}$). Transmission then collapses at the horizon: at $t=10$ it is 0.59 and 0.42, and at $t=20$ it is 0.41 for $\pi_0$ and 0.01 (95\% interval $[-0.02,0.04]$) for $\pi_{0.5}$, whose action no longer moves toward the new task at all; through approach and grasp it stays between 0.36 and 0.49 for $\pi_0$ and recovers only gradually for $\pi_{0.5}$ (0.13 at $t=30$, 0.45 at $t=60$). The attention to the instruction does not follow: $\alpha_L$ remains between 0.92 and 1.20 of its initial value over the same window. The instruction is therefore still read but no longer acted upon (Fig.~\ref{fig:overview}, top)---absorbed downstream of attention, which suggests that boosting the attention to $\lnew$ addresses the wrong quantity (see ``Training-free remedies and ablations''). To test (P1), a linear probe~\cite{probe} predicts the task from instruction-free backbone features (image tokens computed with an empty instruction, mean-pooled per camera). On the spatial pair the tasks are indistinguishable at $t=0$, where the scenes are identical; at $t=10$ the probe already identifies the task with 96\% ($\pi_0$) and 100\% ($\pi_{0.5}$) accuracy and stays above 93\% up to $t=100$. The task becomes visible without language exactly at the third query, the behavioral horizon, and the transmission ratio collapses at the same moment; on LIBERO-Goal (chance 10\%), accuracy reaches 69\% and 84\% at $t=20$, the step at which switches to a different object have collapsed. Opening the policy along a single rollout tells the same story at the level of one state (Fig.~\ref{fig:inside}(a)--(c)): at $t=30$, ``cabinet'' is still the most attended word of $\lnew$ (89\% of its attention at $t=0$), yet $\lnew$ changes only 6\% of the attention over the third-person image (16\% at $t=0$), and the 16 chunks sampled under either instruction all enter the stove bowl.

\subsection*{Main results}
\begin{figure*}[!tp]
\centering
\includegraphics[width=0.85\textwidth]{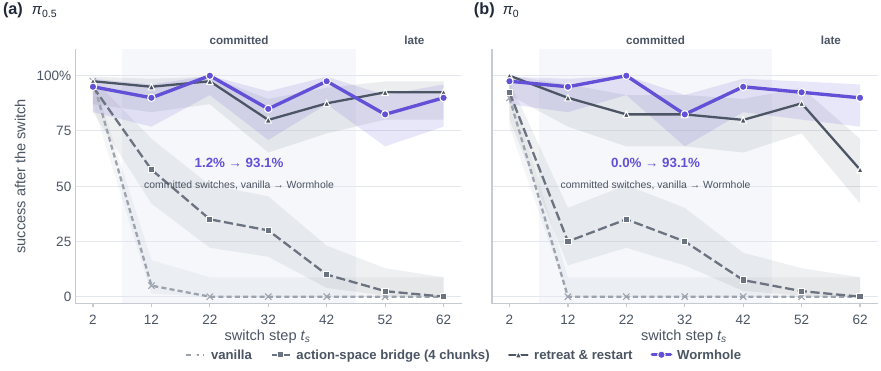}
\caption{\textbf{Success after the switch by switch step} on the LIBERO-Spatial pair (test split, both directions pooled, 40 trials per point, 95\% Wilson bands) for the vanilla policy, the action-space bridge with four chunks, retreat-and-restart, and \ours{}; left $\pi_{0.5}$, right $\pi_0$. The shaded region marks the committed regime; the callout gives the committed-switch success of the vanilla policy and of \ours{} (Table~\ref{tab:main}).}
\label{fig:wormhole}
\end{figure*}

\begin{figure*}[!tp]
\centering
\includegraphics[width=\textwidth]{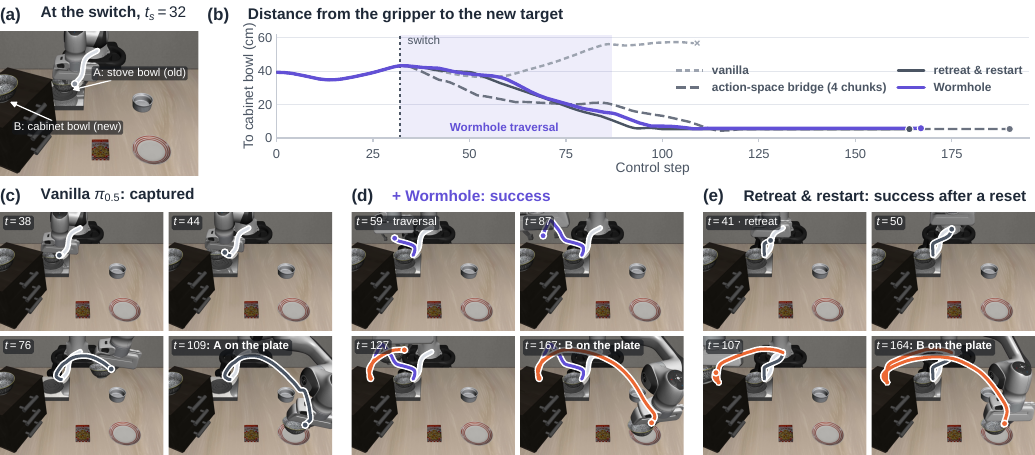}
\caption{\textbf{One switch, three responses.} A paired example on the LIBERO-Spatial pair with $\pi_{0.5}$ (initial state 0), replayed in the simulator with the end-effector path projected into the agent-view camera. All rollouts share the seeded noise and coincide until the instruction is switched from A (the bowl on the stove) to B (the bowl on the wooden cabinet) at $t_s=32$, as the gripper closes on the stove bowl (light path). (a) The shared state at the switch. (b) Distance from the end effector to the cabinet bowl; the shaded band marks the traversal of \ours{}. (c) The vanilla policy puts the stove bowl on the plate (capture). (d) \ours{} opens the gripper, lifts, transits, and settles at the far mouth in the approach phase of B (violet, 55 steps); the unmodified policy, queried under B, then grasps the cabinet bowl and places it on the plate (orange). (e) Retreat-and-restart returns to the initial pose with a scripted controller (dark gray) and restarts under B. The action-space bridge with four chunks (dashed in (b)) also succeeds here, after a longer detour.}
\label{fig:qualitative}
\end{figure*}

Table~\ref{tab:main} and Fig.~\ref{fig:wormhole} compare \ours{} with the vanilla policy, the action-space bridge, and retreat-and-restart from the same initial states. Neither vanilla backbone escapes: $\pi_0$ completes none of its 160 committed switches and $\pi_{0.5}$ two, and 49.4\% and 71.9\% of the trials are captured. Steering the motion without relocating the robot helps only partially: the action-space bridge reaches 14.4--23.1\% on $\pi_0$ and 30.0--33.1\% on $\pi_{0.5}$ and is still captured in 15.0--34.4\% of the trials. \ours{} completes 93.1\% of the committed switches on both backbones (88.1--96.1\%), is never captured, and leaves the early regime intact (97.5\% and 95.0\%). Figure~\ref{fig:qualitative} follows one committed switch from the same state and noise under the vanilla policy, \ours{}, and retreat-and-restart. Retreat-and-restart, which relocates the robot backward to its initial pose, completes 83.8\% ($\pi_0$) and 90.0\% ($\pi_{0.5}$). Pair by pair, on $\pi_{0.5}$ 14 switches succeed only with \ours{} and 9 only with the retreat (exact McNemar test, $p=0.41$); on $\pi_0$ the counts are 22 and 7 ($p=0.008$) in favor of \ours{}. The two relocations differ in how they move the robot: \ours{} hands the robot back after a median traversal of 43 ($\pi_{0.5}$) and 44 ($\pi_0$) steps and the retreat after 19 and 20, yet both complete the new task about equally fast (138 versus 136 steps on $\pi_{0.5}$, 141 versus 138 on $\pi_0$), because \ours{} delivers the robot into the approach phase rather than to its start. The rate-limited traversal changes the action at the switch by 0.32 and 0.30, one fifth of the jump of the retreat (1.49 and 1.44), with less than a third of its jerk (0.09 versus 0.30 and 0.29), and it needs no home pose. The gain holds in both directions (Fig.~\ref{fig:wormhole}): on $\pi_{0.5}$, \ours{} completes 90.0\% of the committed A$\to$B and 96.2\% of the B$\to$A switches (retreat 100\% and 80.0\%; bridge with four chunks 62.5\% and 3.8\%), on $\pi_0$ 98.8\% and 87.5\% (retreat 90.0\% and 77.5\%). Late switches, issued after the old bowl has been grasped, remain out of reach for the vanilla policy and the bridge (at most 1 of 80), whereas both relocations open the gripper, drop the bowl, and succeed in most: \ours{} in 86.2\% ($\pi_{0.5}$) and 91.2\% ($\pi_0$), the retreat in 92.5\% and 72.5\% ($p=0.30$ and $0.004$).

\begin{table*}[!tp]
\footnotesize
\centering
\caption{\textbf{Mid-execution switching on the LIBERO-Spatial pair} (test split, initial states 0--19, both directions; 40 early, 160 committed, and 80 late trials per row). Early, Committed, and Late are success rates in \% (95\% Wilson interval of the committed rate in brackets); Capture and Freeze are over committed trials; Completion is the median number of steps from the switch to the goal of $\lnew$ over successful committed trials ($^\dagger$: fewer than ten successes); Disc.\ and Jerk measure tidal disruption (Eq.~\eqref{eq:metrics}). Best values per backbone in bold.}
\label{tab:main}
\renewcommand{\arraystretch}{1.2}
\setlength{\tabcolsep}{3pt}
\begin{tabularx}{\textwidth}{@{}l Y{0.84}Y{1.72}Y{0.88}Y{0.84}Y{1.2}Y{0.84}Y{0.84}Y{0.84}@{}}
\toprule
\headrow
Method & Early\,\up & Committed\,\up & Capture\,\down & Freeze\,\down & Completion\,\down & Disc.\,\down & Jerk\,\down & Late\,\up \\
\midrule
$\pi_0$~\cite{pi0} & 90.0 & 0.0 {\scriptsize[0.0, 2.3]} & 49.4 & 50.6 & n/a & \best{0.08} & 0.17 & 0.0 \\
\quad + action-space bridge (2 chunks) & 95.0 & 14.4 {\scriptsize[9.8, 20.6]} & 18.8 & 66.9 & 159 & 0.34 & 0.15 & 1.2 \\
\quad + action-space bridge (4 chunks) & 92.5 & 23.1 {\scriptsize[17.3, 30.2]} & 15.0 & 61.9 & 159 & 0.34 & 0.15 & 1.2 \\
\quad + retreat-and-restart (scripted reset) & \best{100.0} & 83.8 {\scriptsize[77.3, 88.7]} & \best{0.0} & 16.2 & \best{138} & 1.44 & 0.29 & 72.5 \\
\oursrow \quad + \ours{} & 97.5 & \best{93.1 {\scriptsize[88.1, 96.1]}} & \best{0.0} & \best{6.9} & 141 & 0.30 & \best{0.09} & \best{91.2} \\
\addlinespace[3pt]
$\pi_{0.5}$~\cite{pi05} & 97.5 & 1.2 {\scriptsize[0.3, 4.4]} & 71.9 & 26.9 & 120$^\dagger$ & \best{0.09} & 0.12 & 0.0 \\
\quad + action-space bridge (2 chunks) & \best{100.0} & 30.0 {\scriptsize[23.4, 37.5]} & 34.4 & 35.6 & 147 & 0.32 & 0.16 & 1.2 \\
\quad + action-space bridge (4 chunks) & 95.0 & 33.1 {\scriptsize[26.3, 40.7]} & 31.2 & 35.6 & 148 & 0.32 & 0.16 & 1.2 \\
\quad + retreat-and-restart (scripted reset) & 97.5 & 90.0 {\scriptsize[84.4, 93.8]} & \best{0.0} & 10.0 & \best{136} & 1.49 & 0.30 & \best{92.5} \\
\oursrow \quad + \ours{} & 95.0 & \best{93.1 {\scriptsize[88.1, 96.1]}} & \best{0.0} & \best{6.9} & 138 & 0.32 & \best{0.09} & 86.2 \\
\bottomrule
\end{tabularx}
\end{table*}

\noindent\textbf{Relocation decides the escape.} Prediction (P3) states that success is decided by where the robot is handed back. Across every method and every committed switch of the spatial pair (1360 trials for $\pi_{0.5}$, 960 for $\pi_0$), we measure the distance from the end effector at hand-over to the approach phase of successful executions of $\lnew$ (Supplementary Note~2). Success falls steeply with this distance, almost identically for both backbones: hand-overs within 3\,cm succeed in 87.7\% ($\pi_{0.5}$) and 84.3\% ($\pi_0$) of the trials, those beyond 15\,cm in 3.4\% and 2.5\%. The methods separate along the same axis: inference-time remedies leave the robot where the vanilla policy is, a median of 20\,cm from the new task, the action-space bridge brings it to 9--18\,cm, and only the two relocations hand over within 3\,cm (\ours{} in 98.1\% and 98.8\% of its trials, the retreat in 99.4\% and 100\%). The bridge fails for exactly this reason: it transfers how the entries of the new task move, but not where they are. In the rollout of Fig.~\ref{fig:teaser}(b), the far mouth chosen online lies 29 steps into an execution of B with 84 steps to go, and it coincides to within 0.1\,mm with the $b$-weighted entry, as expected when the exits form a single mode (Methods).

\noindent\textbf{When \ours{} fails.} \ours{} fails in 11 of 160 committed switches on each backbone, never by capture. In three ($\pi_{0.5}$) and two ($\pi_0$) of them the traversal does not arrive within its budget of 90 steps (the arm, raised above the cabinet, tracks the transit at a fraction of the commanded speed and ends 21--25\,cm short); in one trial of $\pi_{0.5}$ the gate found the robot within 2\,cm of the approach phase of B and handed the instruction directly to the policy, which froze; the remaining seven and nine freeze after the robot has been delivered within 2\,cm of the far mouth, eight of the nine on $\pi_0$ in the B$\to$A direction, where the retreat also freezes most often (22.5\%).

\noindent\textbf{LIBERO-Goal.} Table~\ref{tab:goal} repeats the comparison on LIBERO-Goal with an unchanged configuration and memories built from held-out no-switch rollouts (initial states 20--29). Switches that only change the destination are already followed by the vanilla policy, and \ours{} preserves this (100\%). For switches to a different object, \ours{} raises success from 25.0\% and 14.6\% to 70.8\% and 91.7\% on $\pi_{0.5}$ at $t_s=20$ and 40, and from 4.2\% and 0\% to 64.6\% and 79.2\% on $\pi_0$, again without a single capture; at $t_s=40$ it matches or exceeds retreat-and-restart (89.6\% and 70.8\%), but at $t_s=20$ it falls short of it (97.9\% and 89.6\%). Three of the six pairs account for 17 of its 18 failures on $\pi_{0.5}$ and 25 of 27 on $\pi_0$ and expose two weaknesses: for the switch from the stove to the middle drawer, the far mouth lies at the end of a long approach and the traversal stops 6--9\,cm short of it in all 32 trials (the policy still opens the drawer from there in 12 of 16 trials at $t_s=40$ but in 2 of 16 at $t_s=20$); for the switch from the wine bottle to the bowl at $t_s=20$, the early approach paths of the two tasks overlap, so the gate passes the instruction directly to the policy in 14 of 16 trials, which then freezes in 12. Retreat-and-restart avoids both traps, because it needs neither a reachable far mouth nor a gate.

\begin{table*}[!tp]
\footnotesize
\centering
\caption{\textbf{Switching on LIBERO-Goal} (initial states 0--7; retreat: retreat-and-restart). Left: successes in 8 trials for every ordered pair that switches to a different object, at the switch step $t_s$. Middle: success with its 95\% Wilson interval, capture, and freezing over these six pairs (48 trials per row), in \%. Right: successes in 8 trials for the two pairs that move the same bowl to the other place. Best success, capture, and freezing per backbone and switch step in bold.}
\label{tab:goal}
\renewcommand{\arraystretch}{1.12}
\setlength{\tabcolsep}{1.5pt}
\definecolor{tgzero}{gray}{0.6}\newcommand{\tgz}{\textcolor{tgzero}{0}}
\newcommand{\tgpair}[2]{\begin{tabular}[b]{@{}c@{}}#1\\$\to$\,#2\end{tabular}}
\begin{tabularx}{\textwidth}{@{}l >{\hspace{4pt}}c *{6}{Y{0.93}} Y{1.81} Y{0.95} Y{0.80} *{2}{Y{0.93}}@{}}
\toprule
\headrow
 & & \multicolumn{6}{c}{Different object: successes of 8} & \multicolumn{3}{c}{Different object, pooled (\%)} & \multicolumn{2}{c@{}}{Same bowl} \\
\headrow
Method & $t_s$ & \tgpair{drawer}{stove} & \tgpair{stove}{drawer} & \tgpair{cheese}{wine} & \tgpair{wine}{cheese} & \tgpair{bowl}{wine} & \tgpair{wine}{bowl} & Success\,\up & Capture\,\down & Freeze\,\down & \tgpair{stove}{plate} & \tgpair{plate}{stove} \\
\midrule
$\pi_0$~\cite{pi0} & 20 & \tgz & \tgz & 2 & \tgz & \tgz & \tgz & 4.2 {\scriptsize[1.2, 14.0]} & \best{0.0} & 95.8 & 8 & 8 \\
 & 40 & \tgz & \tgz & \tgz & \tgz & \tgz & \tgz & 0.0 {\scriptsize[0.0, 7.4]} & 27.1 & 72.9 & 8 & 8 \\
\quad + retreat & 20 & 8 & 7 & 7 & 6 & 7 & 8 & \best{89.6 {\scriptsize[77.8, 95.5]}} & \best{0.0} & \best{10.4} & 7 & 8 \\
 & 40 & 5 & 7 & 8 & 6 & 8 & \tgz & 70.8 {\scriptsize[56.8, 81.8]} & \best{0.0} & 29.2 & 8 & 7 \\
\oursrow \quad + \ours{} & 20 & 8 & 2 & 4 & 8 & 7 & 2 & 64.6 {\scriptsize[50.4, 76.6]} & \best{0.0} & 35.4 & 8 & 8 \\
\oursrow & 40 & 8 & 5 & 2 & 8 & 7 & 8 & \best{79.2 {\scriptsize[65.7, 88.3]}} & \best{0.0} & \best{20.8} & 8 & 8 \\
\addlinespace[3pt]
$\pi_{0.5}$~\cite{pi05} & 20 & 1 & 4 & \tgz & 4 & 1 & 2 & 25.0 {\scriptsize[14.9, 38.8]} & 2.1 & 72.9 & 8 & 8 \\
 & 40 & \tgz & \tgz & \tgz & 4 & 3 & \tgz & 14.6 {\scriptsize[7.2, 27.2]} & 20.8 & 64.6 & 8 & 8 \\
\quad + retreat & 20 & 8 & 8 & 8 & 8 & 7 & 8 & \best{97.9 {\scriptsize[89.1, 99.6]}} & \best{0.0} & \best{2.1} & 8 & 8 \\
 & 40 & 8 & 8 & 8 & 8 & 6 & 5 & 89.6 {\scriptsize[77.8, 95.5]} & \best{0.0} & 10.4 & 8 & 8 \\
\oursrow \quad + \ours{} & 20 & 8 & \tgz & 8 & 8 & 8 & 2 & 70.8 {\scriptsize[56.8, 81.8]} & \best{0.0} & 29.2 & 8 & 8 \\
\oursrow & 40 & 8 & 7 & 8 & 8 & 7 & 6 & \best{91.7 {\scriptsize[80.4, 96.7]}} & \best{0.0} & \best{8.3} & 8 & 8 \\
\bottomrule
\end{tabularx}
\end{table*}

\noindent\textbf{Overhead.} \ours{} adds computation once per switch: sampling the $N=16$ exits (901\,ms on an idle NVIDIA L20 with the $\pi_{0.5}$ server, against 102\,ms for an ordinary query) and selecting the far mouth (about 10\,ms on one CPU core for the 2441 states of the memory). During the traversal the policy is not queried, which saves about nine queries over the median traversal; on a robot, the arm holds its pose during that second.

\subsection*{Training-free remedies and ablations}
Prediction (P2) states that inference-time fixes can revive a weak instruction signal but cannot create one that was never learned. We test it on $\pi_{0.5}$ (initial states 0--9; 20 early and 80 committed trials per method; full tables in Supplementary Note~3) with three interventions that leave the policy untouched: \emph{flush} discards the queued actions and re-queries the policy at $t_s$; \emph{attention re-routing} adds a bias $\beta$ to the attention logits from the action queries to the tokens of $\lnew$~\cite{pasta}; and \emph{contrastive instruction guidance} (CIG) extrapolates away from the old instruction, $\tilde v=v_\theta(z,\tau\mid\cnew)+w\,[v_\theta(z,\tau\mid\cnew)-v_\theta(z,\tau\mid\cold)]$, in the spirit of classifier-free guidance and contrastive decoding~\cite{cfg,contrastive}. Flushing the queue and re-routing attention ($\beta=2$, 5) raise committed success only from 1.2\% to 3.8--5.0\% and leave capture at 47.5--71.2\%: the black hole is neither execution latency nor a lack of attention to $\lnew$. CIG helps more (11.2\% for $w=1$ and $w=3$), but exactly as (P2) predicts: it succeeds in 8--9 of 20 trials at $t_s=12$, right after the horizon, and in at most 1 of 20 at every later step, while degrading the early regime (85\% and 70\% instead of 95\%) and raising the switch discontinuity to 0.56--0.76. On the same trials the action-space bridge reaches 32.5\%, retreat-and-restart 88.8\%, and \ours{} 95.0\%; of the two relocations only \ours{} keeps the action stream continuous (discontinuity 0.32 versus 1.51).

Ablating \ours{} on the same trials shows that the \emph{phase} of the entries is the component that matters: when entries may also come from after the grasp, the far mouth lands where the executions of $\lnew$ carry their object, and the policy, delivered there with an empty gripper, freezes in 98.8\% of the switches. Restricted to the approach phase, the far mouth is robust: a phase margin of $m=20$ (96.2\%), the nearest entry without time-to-go (97.5\%), and the $b$-weighted entry without exits and OT (92.5\%) all lie within the interval of the full method (95.0\%). Over the 315 committed traversals of the test split and the 184 on LIBERO-Goal, the far mouth chosen with OT lies within 0.5\,mm of the $b$-weighted entry, as expected when the exits form a single mode: what makes \ours{} work here is where it relocates the robot, not how the plan is computed. The traversal trades smoothness for time: without the rate limit, success is unchanged (93.8\%) but the switch discontinuity rises from 0.32 to 1.50, as large as the retreat's; without the lift and clearance, \ours{} succeeds as often (97.5\%) and finishes 20 steps sooner, because in these scenes the direct transit collides with nothing, and we keep the lift as a conservative default. The action-space bridge, which shares the mouths and the throat, stays at 16.2--32.5\% whatever its length, bridge field, or learned bridge head.

\subsection*{Distilling the wormhole into the policy}
\phantomsection\label{sec:distill}%
\ours{} leaves the policy deaf and moves the robot instead. Remark~\ref{rem:predictions} predicts the alternative: switching data, which make committed states ambiguous, should let a policy learn to listen. We test this and ask whether the instruction can be routed through an explicit grounding pathway, as in vision-language models: at the switch moments of the development split, the box-coordinate position of Qwen2.5-VL-7B moves its attention to the newly referred bowl (pointing rate 0.65), whereas the instruction position of $\pi_{0.5}$ stays on the old one (pointing rate 0.22; Fig.~\ref{fig:inside}(d)).

\noindent\textbf{Attention wormhole.} We append a learnable grounding token [LOC] to the prefix of $\pi_{0.5}$. It attends to both camera views and the instruction, while the other prefix tokens do not attend to it. In heads 0--3 of layers 6--13, its attention over the $16\times16$ base-camera patches is supervised to match a Gaussian ($\sigma=14$\,px) at the projection of the referred object, and a readout head predicts the object's pixel position and its offset from the gripper. In the action expert, [LOC] contributes its key and value, and an attention bias adds a gated, clipped log-map of its grounding to the logits of the image patches. In the attention wormhole (AW), the image tokens are computed without the instruction and the action expert does not attend to the instruction tokens, so the instruction reaches the actions only through [LOC]; a leak test with instructions of different lengths confirms this. Two controls serve as references: AW-words, whose action expert may also read the words, and AW-bias, which keeps only the attention bias.

\noindent\textbf{Switching data without human demonstrations.} The training data are generated by an oracle relocation teacher: at the switch, a scripted traversal carries the arm 15\,cm above the grasp point of the new bowl, read from the simulator, and the released $\pi_{0.5}$ completes the task. We keep the 596 successful teacher episodes from initial states 20--39 (477 with a switch in either direction and 119 without; 98{,}894 frames, initial states 38--39 held out for validation) and train LoRA adapters of both experts together with [LOC], its readout and the bias gates (53.2M of 3404.5M parameters) for 6000 steps with the flow-matching loss plus 0.1 times the grounding KL and 0.2 times the readout loss. The behavior-cloning (BC) baseline uses the same data, adapters, and schedule without [LOC]. Geometric augmentation of the base camera (random crop 95--100\%, rotation $\pm5^\circ$) and one round of DAgger data were compared on the development split before any test run; augmentation was selected (94.2\% versus 88.3\%; DAgger 81.7\%), the selected configuration and a BC baseline with the same augmentation were trained twice, and every model was evaluated once on the test split (Table~\ref{tab:distill}).

\begin{table*}[!tp]
\footnotesize
\centering
\caption{\textbf{Self-trained policies on the LIBERO-Spatial pair} ($\pi_{0.5}$, test split, initial states 0--19, both directions; 40 early, 160 committed, and 80 late trials per training run, pooled over the runs of a row; columns as in Table~\ref{tab:main}). All self-trained rows use the switching data of the oracle relocation teacher; every model is evaluated once. AW: attention wormhole with the instruction routed only through [LOC]; aug.: geometric augmentation; safe exit: teacher data whose relocation hands back before the elbow approaches its joint limit.}
\label{tab:distill}
\renewcommand{\arraystretch}{1.2}
\setlength{\tabcolsep}{3pt}
\begin{tabularx}{\textwidth}{@{}l Y{1.9}Y{0.62}Y{1.46}Y{0.74}Y{0.62}Y{1.04}Y{0.62}@{}}
\toprule
\headrow
Method & Training & Early\,\up & Committed\,\up & Capture\,\down & Freeze\,\down & Completion\,\down & Late\,\up \\
\midrule
$\pi_{0.5}$~\cite{pi05} & none & 97.5 & 1.2 {\scriptsize[0.3, 4.4]} & 71.9 & 26.9 & 120$^\dagger$ & 0.0 \\
\oursrow \quad + \ours{} (plug-in) & none & 95.0 & \best{93.1 {\scriptsize[88.1, 96.1]}} & \best{0.0} & \best{6.9} & \best{138} & \best{86.2} \\
\addlinespace[3pt]
BC & switching data & 97.5 & 87.5 {\scriptsize[81.5, 91.8]} & \best{0.0} & 12.5 & 151 & 71.2 \\
\oursrow AW (instruction only via [LOC]) & switching data & \best{100.0} & 89.4 {\scriptsize[83.6, 93.3]} & \best{0.0} & 10.6 & 148 & 70.0 \\
BC + aug.\ (2 runs) & switching data & 97.5 & 88.8 {\scriptsize[84.8, 91.8]} & \best{0.0} & 11.2 & 150 & 66.9 \\
\oursrow AW + aug.\ (2 runs; selected on dev.) & switching data & 98.8 & 90.9 {\scriptsize[87.3, 93.6]} & \best{0.0} & 9.1 & 147 & 73.1 \\
\addlinespace[3pt]
BC + aug.\ (3 runs) & safe-exit switching data & 98.3 & 85.2 {\scriptsize[81.8, 88.1]} & \best{0.0} & 14.8 & 149 & 63.8 \\
\oursrow AW + aug.\ (3 runs) & safe-exit switching data & 97.5 & 92.1 {\scriptsize[89.3, 94.2]} & \best{0.0} & 7.9 & 149 & 76.7 \\
\bottomrule
\end{tabularx}
\end{table*}

\noindent\textbf{Switching data close the black hole.} Trained on the teacher's switches, both policies follow new instructions inside the committed region, where the released policy follows 1.2\%: BC reaches 87.5\% and AW 89.4\%, as the remark predicts. Routing the instruction through [LOC] costs no accuracy. With augmentation and pooled over two runs, AW succeeds in 90.9\% of committed and 73.1\% of late switches against 88.8\% and 66.9\% for BC; in the pre-registered comparison over committed and late trials, paired within each run, AW succeeds in 408 and BC in 391 of 480 trials, 3.5 points (95\% CI $[-0.3,7.4]$, exact McNemar $p=0.089$): not worse than BC, but not significantly better. Three diagnostics on the development split show that the correction is carried by [LOC] (Supplementary Note~4): [LOC] points to the referred bowl at all 100 switch moments, whereas the best heads of BC point to it in 0.31--0.42 (Fig.~\ref{fig:inside}(d)); hiding [LOC] from the action expert at the switch drops committed success from 95.0\% to 0.0\% (and from 91--96\% to 5--20\% for the safe-exit models, to 0\% in every B$\to$A trial), whereas removing only the attention bias changes nothing; and when the action expert may also read the words (AW-words), it ignores the wormhole---it still succeeds in 93.8\% with [LOC] and the bias switched off, and pairing the old words with the new [LOC] sends every trial to the old bowl. Routing is therefore what turns a grounding token into a causal pathway.

\noindent\textbf{A kinematically safe exit.} The failures that remain after training are almost all B$\to$A switches that freeze, and they have a mechanical signature: replaying the recorded trajectories shows the elbow joint at its upper limit (fully extended) at the switch in 35--55\% of the frozen and 87--100\% of the captured trials, and in none of the successful trials or teacher episodes. The arm is straightened while the student still executes the old task---reaching for the far cabinet bowl from above, 73--75\,cm from the base where the teacher stays at 68\,cm---and from that singular configuration no controller we tried returns in time. The posture is inherited from the teacher: the only episodes in which the oracle relocation brings the arm high, far, and nearly straight are the A$\to$B switches, whose hand-back point lies 15\,cm above the far bowl (elbow margin 0.86\,rad, against 1.2--1.6\,rad when the released policy reaches the same bowl by itself). What helps is to move the \emph{exit} of the wormhole: the relocation hands the arm back as soon as the elbow margin falls below 1.2\,rad (selected on a pilot of 80 teacher episodes), and the policy completes the approach from a dexterous posture; B$\to$A relocations, which fold the arm, are unaffected. The new teacher succeeds in 243 of the 260 A$\to$B conditions (the original in 250), giving 589 episodes and 97{,}397 frames. AW and BC were retrained twice on these data; the development split favored the retrained AW (638 versus 629 of 720 trials), the test split was run once per model, and a third run of both was trained as a confirmation and tested once, whatever its development result. Pooled over three runs, AW succeeds in 92.1\% of committed and 76.7\% of late test switches against 85.2\% and 63.8\% for BC. In the pre-registered comparison over the first two runs, AW succeeds in 419 and BC in 372 of 480 trials, 9.8 points (95\% CI $[5.9,13.7]$, $p=1.5\times10^{-6}$); the third run reproduces it (207 versus 190, $p=0.009$), each run is significant on its own ($p=8\times10^{-6}$, $0.015$, $0.009$), and the three pooled give 626 versus 562 of 720, 8.9 points (95\% CI $[5.8,12.0]$, $p=3\times10^{-8}$), concentrated in B$\to$A (80.3\% versus 63.6\%) and late switches. Two comparisons qualify this result: against the policies trained on the original data, the safe-exit data improve AW by 2.3 points (419 versus 408, $p=0.25$) but lower BC by 4.0 points (372 versus 391, $p=0.018$), entirely in B$\to$A switches, and replaying every B$\to$A development trial shows why: the safe-exit data do not change how often AW straightens its elbow before the switch (11 of 80 trials in both runs, against 12 and 8) but make BC do so twice as often (20 and 20 against 12 and 11), and no trial that reached the joint limit before the switch succeeded. Why the two pathways react so differently is open. With no plug-in at test time, the best self-trained policy reaches the training-free \ours{} on committed switches (92.1\% versus 93.1\%) and approaches it on late ones (76.7\% versus 86.2\%).

\noindent\textbf{Second backbone.} The configuration does not transfer to $\pi_0$. Trained in the same way on switching data whose policy segments come from the released $\pi_0$ (whose teacher succeeds in 63.1\% of its episodes, against 91.7\% with $\pi_{0.5}$), AW follows 46.2\% of committed and 43.8\% of late test switches against 62.5\% and 52.5\% for BC (109 versus 142 of 240 paired trials, $p<0.001$), although [LOC] is learned as well as on $\pi_{0.5}$ (pointing rate 1.0): the deficit lies in how the action expert of $\pi_0$ uses it. The safe-exit data narrow this gap without closing it: with the same elbow-margin exit (new teacher 226 of 260 A$\to$B conditions, original 220), two runs each, AW follows 55.9\% of committed and 41.9\% of late switches against 62.5\% and 49.4\% for BC (246 versus 279 of 480, $p=0.006$). The data raise AW by 5.8 points ($p=0.018$)---its A$\to$B switches rise from 64.2\% to 83.3\%, on par with BC, while its B$\to$A switches fall from 26.7\% to 19.2\% ($p=0.03$)---and leave BC unchanged (59.2\% versus 58.1\%, $p=0.70$). We therefore report the self-trained wormhole as validated on $\pi_{0.5}$ only; the training-free \ours{} is unaffected.

\section*{Discussion}

We identified the instruction black hole: after a VLA policy commits to a task, new instructions are absorbed without effect. The black hole is a blind spot of static imitation rather than a limitation of capacity, and the same blind spot locks a policy onto any task whose committed region it is placed in. \ours{} turns this cause into a cure: it chooses a far mouth by OT between the policy's sampled continuations and approach-phase states of successful executions of the new task, carries the robot there with a rate-limited traversal, and lets the unmodified policy finish. The horizon is abrupt and precedes physical commitment, and the instruction stops being transmitted at the moment a language-free probe can read the task from the state, while the attention to it stays intact. Without any training, \ours{} raises committed-switch success from at most 1.2\% to 93.1\% on both backbones, is never captured, and matches or exceeds a scripted retreat-and-restart with one fifth of its discontinuity; and the switching data it generates let a policy learn to listen by itself.

Every method that reliably escapes the black hole moves the robot to a state from which the policy executes the new task, and the others succeed only when the robot happens to end up there. The scripted retreat relocates backward, to the initial pose before the horizon, where the task is not yet identifiable and the instruction is heard again; \ours{} relocates forward, into the approach phase of the new task, where the state already identifies it. Rather than making a policy listen where static data taught it not to, one can choose where to give it the floor. On the spatial pair the forward relocation is at least as reliable as the backward one, smoother, and needs no home pose; the backward one needs no memory of the new task and was more reliable on LIBERO-Goal at $t_s=20$, where the far mouth of the drawer pair is not reached within the traversal budget and the position-only gate hands the instruction to a policy that is still committed. A gate that tests task identifiability (for example with the language-free probe of Fig.~\ref{fig:mechanism}(a)) and a planned traversal path would address both, and the two relocations compose: when no far mouth can be reached, retreating remains a fallback. After the grasp, both relocations release the held object implicitly and succeed in most late switches, which LIBERO tolerates; on a real robot, late switches call for an explicit place-or-release step.

Remark~\ref{rem:predictions} predicts that the blind spot is a property of the data rather than of the model, and Table~\ref{tab:distill} confirms it: switching data generated by the relocation teacher close the black hole even for plain behavior cloning, and routing the instruction through an explicit grounding token makes the correction traceable to one token without costing accuracy. Training-free relocation needs no data and no access to the weights; the self-trained policy needs both but no plug-in at test time, and the two compose, because the plug-in generates the switching data from which the policy learns. What remains of the black hole after training is not in the attention---[LOC] still points to the new bowl at the frozen frames---but in the joint configuration, which the policy does not receive as input: the student straightens the elbow while executing the old task, a posture inherited from the teacher's hand-back point, and from the joint limit no controller returns in time. The exit of a wormhole therefore matters as much as its entry: it must lie inside the dexterous workspace, not merely above the target.

By Proposition~\ref{prop:blind}, a policy can achieve perfect static success while being deaf on most of its trajectory, so static benchmarks systematically overestimate how well VLA policies use language; we advocate reporting escape curves $\Pesc(t_s)$ alongside static success rates. The black hole is not specific to language: any conditioning signal that becomes redundant once execution is under way---goal images, user preferences, safety constraints---can be absorbed in the same way, and relocation may help there as well.

Our experiments are in simulation, on LIBERO, with two flow-based backbones of the same family; autoregressive policies are not evaluated, and validation on a real robot with spoken interruptions is the natural next step. \ours{} needs a memory of successful executions of the new task in the same workspace, and its far mouth is an absolute end-effector pose, so large displacements of the objects call for entries expressed relative to the objects. The traversal follows a fixed lift--transit--settle path that assumes free space above the workspace, can stall or stop short of a far mouth that is hard to reach, and its position-only gate can skip a traversal that was needed; late switches release the grasped object. On our scenes the OT coupling did not change the far mouth, so its benefit for multimodal continuations remains to be shown, and Proposition~\ref{prop:blind} assumes a sharply defined committed region, whereas real data produce a soft horizon that the transmission ratio measures but the analysis does not quantify. The self-trained attention wormhole is evaluated on one pair with two or three training runs per configuration; the safe-exit data raise it by 2.3 points (not significant) but lower behavior cloning by 4.0 points (significant), so part of the gap between the two safe-exit policies comes from behavior cloning getting worse, and the safe-exit teacher itself succeeds in 243 instead of 250 of 260 conditions. On $\pi_0$ the self-trained wormhole stays below behavior cloning with both data sets, and the teacher uses the simulator's object positions. Five further attempts to improve the trained policy (self-imitation from switch self-play, contrastive guidance at inference, and three relabeling schemes after the switch) did not raise development success and were not tested.

We hope that the black hole lens will help the community evaluate, and build, robots that listen not only when they start but also while they act.

\section*{Methods}

\subsection*{Wormhole mouths}
\noindent\textbf{Mouth A: old-task exits.} Under $\cold$ we draw $N$ short-horizon continuations from the base policy,
\begingroup\predisplaypenalty=0 % npj layout: a column break may fall just before Eq. (6)
\begin{equation}
\label{eq:pold}
\begin{gathered}
X_{\rm old}=\{\xold_i\}_{i=1}^{N},\qquad \xold_i\sim\pi_\theta(\cdot\mid\cold),\\
\pold=\sum_{i=1}^{N}a_i\,\delta_{\xold_i},
\end{gathered}
\end{equation}
\endgroup
with uniform weights $a_i$; they describe where the black hole is pulling the robot, and $q_i$ denotes the end-effector position at which exit $i$ would arrive.

\noindent\textbf{Mouth B: new-task entries.} A memory $\mathcal{M}_{\rm new}$ stores successful executions of $\lnew$ (demonstrations or rollouts). Its entries are the states $y_j$ (end-effector position $p_j$ and orientation) visited during the \emph{approach phase}, at least $m$ steps before the gripper first closes (or, without a grasp, before the task's object first moves), so that the policy still performs the whole manipulation from them; each carries its time-to-go $T_j-k_j$, the steps its execution still needed. The entries are weighted by the expected time to complete $\lnew$ through them,
\begin{equation}
\label{eq:pnew}
\begin{gathered}
\pnew=\sum_{j=1}^{M}b_j\,\delta_{y_j},\qquad b_j\propto\exp\!\big(-E(\chi_{t_s},y_j)/\kappa\big),\\
E(\chi,y_j)=\frac{L(\chi,y_j)}{\bar v}+\lambda\,(T_j-k_j),
\end{gathered}
\end{equation}
where $L$ is the length of the traversal path from the current position to $p_j$ (see ``Traversal and relocation''), $\bar v$ a nominal speed, $\lambda$ trades distance against progress, $\kappa$ is a temperature, and the $M$ lowest-energy entries are retained. The energy balances two ways of wasting time: entries near the robot lie at the start of the new task, so going there undoes the motion, whereas entries near the new object are far away.

\subsection*{Wormhole throat via optimal transport}
We couple the two mouths by entropic OT~\cite{cuturi,peyre},
\begin{equation}
\label{eq:ot}
\begin{gathered}
\gamma^*=\arg\min_{\gamma\in\Pi(a,b)}\ \sum_{i=1}^{N}\sum_{j=1}^{M}\gamma_{ij}\,C_{ij}-\eta H(\gamma),\\
C_{ij}=\frac{L(q_i,y_j)}{\bar v}+\lambda\,(T_j-k_j),
\end{gathered}
\end{equation}
where $\Pi(a,b)$ is the set of couplings with marginals $a$ and $b$ and $H$ the entropy; the cost is median-normalized and the problem is solved with a few log-domain Sinkhorn iterations. Each exit is matched with the entries that complete the new task fastest from where that exit would bring the robot. The plan induces the far mouth, the transport-weighted entry anchored at the mean exit $\hat x=\sum_i a_i\xold_i$, and the wormhole axis that leads to it:
\begin{equation}
\label{eq:axis}
\begin{gathered}
y^*=\sum_{i,j}\frac{\omega_i}{a_i}\,\gamma^*_{ij}\,y_j,\qquad \dOT_{t_s}=y^*-\chi_{t_s},\\
\omega_i\propto a_i\exp\!\big(-\|\xold_i-\hat x\|^2/\sigma^2\big).
\end{gathered}
\end{equation}
Orientations are averaged as quaternions. The axis is the instruction transduced into geometry: it depends on $\lnew$ only through the entries and the plan, and it lives in the state space in which the commitment lives. When the exits form a single narrow mode, as deep in the black hole, $\gamma^*$ approaches the product of its marginals and $y^*$ the $b$-weighted entry; when the policy is torn between continuations, the plan routes each to its own entries.

\subsection*{Traversal and relocation}
\ours{} traverses the throat along a path that lifts the gripper clear of the old object by $h$, transits above the workspace at the higher of the current and target heights plus a clearance $c$, and settles onto $y^*$; $L$ in Eqs.~\eqref{eq:pnew} and~\eqref{eq:ot} is the length of this path. The path is tracked in closed loop through the policy's own action interface (delta end-effector pose), the gripper is opened as in every entry, and the change of the pose action between consecutive steps is bounded by $r$. This rate limit is the tidal regularizer of \ours{}: it bounds the switch discontinuity and the transition jerk by construction. When the end effector is within $\epsilon$ of $y^*$, control returns to the base policy under $\cnew$; if the robot is already within $\epsilon$ of the approach phase of $\lnew$ when the switch arrives, no traversal is needed and the policy simply receives $\lnew$.

\begin{proposition}[Relocation]\label{prop:reloc}
Let $\Ccal$ and $g$ be as in Definition~\ref{def:committed}, and let $\pi$ minimize $J_\Dcal$ with a loss whose minimizer is the conditional expert distribution (negative log-likelihood or conditional flow matching). Then (i) $\pi(\cdot\mid\chi,l)=\pi^{\rm E}(\cdot\mid\chi,l)$ for $\Dcal$-almost every $\chi\in\Ccal$ with $l=g(\chi)$; and (ii) the deafened minimizer $\tilde\pi$ of Proposition~\ref{prop:blind} satisfies $\tilde\pi(\cdot\mid\chi,l)=\pi^{\rm E}(\cdot\mid\chi,g(\chi))$ for every $l$. Hence, if the traversal delivers the robot to a state $y^*\in\Ccal$ with $g(y^*)=\lnew$, the base policy continues the expert behavior of $\lnew$ whether or not it reads the instruction.
\end{proposition}

Proposition~\ref{prop:reloc} (proof in Supplementary Note~1) is the other side of the black hole: the commitment that absorbed $\lnew$ locks the robot onto it after the traversal, so \ours{} relocates, rather than fights, the black hole. The scripted retreat of Table~\ref{tab:main} exploits the same principle backward, returning the robot before the horizon, where the task is not yet identifiable and the instruction is heard again; \ours{} moves forward into the basin of the new task, without a reset pose and without undoing progress toward the new goal.

\subsection*{Implementation details}
Table~\ref{tab:hyper} lists the configuration of \ours{} and of the evaluation, and Algorithm~\ref{alg:escape} the escape procedure.

\begin{table*}[!tp]
\footnotesize
\centering
\caption{\textbf{Implementation details of \ours{} and of the evaluation.} The configuration was fixed on a development split of initial states before the test runs (see ``Experimental setup'').}
\label{tab:hyper}
\renewcommand{\arraystretch}{1.2}
\begin{tabularx}{\textwidth}{l l X}
\toprule
\headrow
Symbol & Meaning & Value\\
\midrule
$N$ & exits sampled under $\lold$ (horizon) & 16 (10 control steps)\\
$q_i$ & end point of an exit & $1.21$\,cm per action unit and step (least squares on executed rollouts)\\
$m$ & phase margin of the entries & 10 steps before the gripper first closes\\
$M$ & entries retained for the throat & 64\\
$\bar v,\lambda,\kappa$ & speed, time-to-go weight, and temperature of Eq.~\eqref{eq:pnew} & 1.5\,cm/step, 1, 5 steps\\
$\eta$ & entropic regularization (median-normalized cost) & 0.1, 300 log-domain Sinkhorn iterations\\
$\sigma^2$ & anchoring bandwidth & $2\times$ median squared exit-to-mean distance\\
$h$, $c$ & lift above the start, transit clearance & 5\,cm, 3\,cm\\
-- & tracking & $(\text{waypoint}-p)/5$\,cm clipped to $\pm1$; rotation error$/0.5$\,rad clipped to $\pm0.5$\\
$r$ & rate limit of the pose action & 0.3 per step (norm of the 6-D change)\\
$\epsilon$ & arrival and gate radius & 2\,cm; at most 90 traversal steps\\
-- & entry memory & successful no-switch rollouts of $\lnew$, initial states 20--39 (goal: 20--29)\\
-- & action-space bridge & 32 segments, cost weights 1, 0.5, 0.5, 0.2; 2/4/6 chunks of 10 steps, 3-step ramp\\
-- & learned bridge head & MLP $3\times512$, SiLU; $2\times10^4$ AdamW steps, lr $3\times10^{-4}$\\
-- & development split & initial states 40--49 (spatial pair)\\
-- & evaluation & initial states 0--19 (spatial), 0--7 (goal), 0--9 (object; ablations)\\
-- & seeds & simulator seed 7; flow noise seeded by initial state and query index\\
\bottomrule
\end{tabularx}
\end{table*}

\begin{algorithm*}[!t]
\caption{\ours{} escape at an instruction switch}
\label{alg:escape}
\footnotesize
\begin{algorithmic}[1]
\REQUIRE base policy $\pi_\theta$; memory $\mathcal{M}_{\rm new}$ of successful executions of $\lnew$; state $\chi_{t_s}$; instructions $\lold,\lnew$; sizes $N,M$
\IF{$\min_{y}\|p(\chi_{t_s})-p(y)\|<\epsilon$ over the approach phase of $\mathcal{M}_{\rm new}$}
\STATE resume $\pi_\theta(\cdot\mid\cnew)$ and stop \hfill (the instruction is still heard)
\ENDIF
\STATE \textit{Mouth A:} $X_{\rm old}\leftarrow\{\xold_i\sim\pi_\theta(\cdot\mid\cold)\}_{i=1}^{N}$, end points $q_i$, $\hat x\leftarrow\sum_i a_i\xold_i$ \hfill Eq.~\eqref{eq:pold}
\STATE \textit{Mouth B:} approach-phase entries $y_j$ of $\mathcal{M}_{\rm new}$, $b_j\propto\exp(-E(\chi_{t_s},y_j)/\kappa)$, keep the best $M$ \hfill Eq.~\eqref{eq:pnew}
\STATE \textit{Throat:} $C_{ij}\leftarrow L(q_i,y_j)/\bar v+\lambda(T_j-k_j)$; $\gamma^*\leftarrow$ \textsc{Sinkhorn}$(a,b,C,\eta)$ \hfill Eq.~\eqref{eq:ot}
\STATE \textit{Far mouth:} $y^*\leftarrow\sum_{ij}(\omega_i/a_i)\gamma^*_{ij}y_j$ \hfill Eq.~\eqref{eq:axis}
\REPEAT
\STATE \textit{Traverse:} track the lift--transit--settle path to $y^*$, gripper open, $\|a_{k}-a_{k-1}\|\le r$
\UNTIL{$\|p(\chi_k)-p(y^*)\|<\epsilon$ or the step budget is spent}
\STATE resume $\pi_\theta(\cdot\mid\cnew)$ \hfill (Proposition~\ref{prop:reloc})
\end{algorithmic}
\end{algorithm*}

\subsection*{Experimental setup}
\noindent\textbf{Protocol.} We implement mid-execution switching on top of the official LIBERO~\cite{libero} evaluation of openpi~\cite{pi0,pi05}: the policy receives the third-person and wrist images ($224\times224$), is queried every five control steps, and acts for at most 220 steps (300 on LIBERO-Goal, 280 on LIBERO-Object) after ten settling steps. At the switch step $t_s$ the instruction sent to the policy is replaced by $\lnew$; in the vanilla setting the at most four queued actions of the current chunk are still executed. Outcomes are read from the BDDL goal predicates: a trial \emph{succeeds} if the goal of $\lnew$ is reached before that of $\lold$, is \emph{captured} if the goal of $\lold$ is reached first, and \emph{freezes} if neither is reached before the step limit; trials whose old task is completed before $t_s$ are excluded. The flow noise of every query is seeded by the initial state and the query index, so rollouts from the same initial state coincide up to $t_s$ for all switch steps and methods, which makes every comparison a paired counterfactual.

\noindent\textbf{Switch pairs and regimes.} Our main testbed is the pair of LIBERO-Spatial tasks that share one layout: two identical black bowls stand on the stove and on the wooden cabinet, and the instructions ``pick up the black bowl on the stove and place it on the plate'' (A) and ``\ldots on the wooden cabinet \ldots'' (B) differ only in which bowl to fetch, so nothing but language separates A from B at the start. We switch in both directions from initial states 0--19. We also use LIBERO-Goal (ten tasks in one kitchen; six ordered pairs that redirect the robot to a different object and two that keep the object and change its destination) and the LIBERO-Object scene with the alphabet soup and the salad dressing. Regimes are defined from the measured dynamics: \emph{early} switches arrive before the behavioral horizon, \emph{committed} switches between the horizon and the grasp of the old object, and \emph{late} switches after the grasp; on the spatial pair, $t_s=2$, $t_s\in\{12,22,32,42\}$, and $t_s\in\{52,62\}$, offset from the query boundaries.

\noindent\textbf{Backbones and configuration.} We use the released LIBERO checkpoints of $\pi_0$~\cite{pi0} and $\pi_{0.5}$~\cite{pi05} (ten flow-integration steps). \ours{} samples $N=16$ exits, keeps the $M=64$ lowest-energy approach-phase entries, solves Eq.~\eqref{eq:ot} with $\eta=0.1$, and tracks the traversal through the policy's delta end-effector actions; all hyperparameters are listed in Table~\ref{tab:hyper}. The entry memory of each direction contains the successful no-switch rollouts of $\lnew$ from the held-out initial states 20--39, so no evaluation episode contributes to it. The configuration was fixed on a development split (initial states 40--49) before any test episode: there, \ours{} succeeded in 95.0\% ($\pi_{0.5}$) and 91.2\% ($\pi_0$) of 80 committed switches with a phase margin of $m=10$ steps and in 95.0\% and 90.0\% with $m=20$, and we kept $m=10$. Every test number of \ours{} comes from a single run on initial states 0--19, in which retreat-and-restart was run again on the same states. We also report the \emph{action-space bridge} of our earlier design, which shares Mouth~A, retrieves 32 state-compatible motion segments of $\lnew$ as Mouth~B and, instead of moving the robot, executes two to six ten-step chunks that shift the mean exit by the transport-weighted motion of the entries. The vanilla, bridge, and inference-time rows come from an earlier run of the same protocol on the same states and seeds; their pre-switch paths agree with the test run to a median deviation of 0.1\,mm ($\pi_{0.5}$) and 1\,mm ($\pi_0$).

\noindent\textbf{Metrics.} \textbf{Early}, \textbf{Committed}, and \textbf{Late} are success rates in the three regimes; \textbf{Capture} and \textbf{Freeze} are the shares of captured and frozen committed trials; \textbf{Completion} is the median number of steps from $t_s$ to the goal of $\lnew$ over successful committed trials; the escape time $\Tesc$ is the number of steps until the end effector has moved 2\,cm closer to the object of $\lnew$ and keeps approaching it. Tidal disruption is measured by the switch discontinuity and the transition jerk of the executed pose actions,
\begin{equation}
\label{eq:metrics}
\begin{gathered}
D_{\rm disc}=\|a_{t_s}-a_{t_s-1}\|_2,\\
J_{\rm trans}=\frac{1}{H-2}\sum_{k=t_s}^{t_s+H-3}\|a_{k+2}-2a_{k+1}+a_k\|_2,\quad H=20 .
\end{gathered}
\end{equation}

\noindent\textbf{Protocol details.} The policy receives the third-person and wrist images rendered at $256\times256$ and resized to $224\times224$. Switch steps on the spatial pair are offset from the query boundaries so that the vanilla policy still executes queued actions of the old chunk. The eight ordered pairs of LIBERO-Goal are: open the middle drawer $\leftrightarrow$ turn on the stove, put the cream cheese in the bowl $\leftrightarrow$ put the wine bottle on top of the cabinet, put the bowl on the plate $\leftrightarrow$ put the wine bottle on the rack (six pairs that redirect the robot to a different object), and put the bowl on the stove $\leftrightarrow$ put the bowl on the plate (two pairs that keep the object and change its destination). On the development split of the spatial pair, ties between configurations are broken by the shorter completion time.

\noindent\textbf{Reproduction.} The $\pi_0$ experiments on the spatial pair were run twice, on two different machines with the same code and seeds. In the first run, 210 trials were interrupted by a failure of the policy server and had to be excluded; the second run completed every trial and is the one reported here. The two runs agree within their 95\% intervals: both give $t_h=10$ in both directions, 90.0\% early success for the vanilla policy, committed success of 1.9\% versus 0.0\% for the vanilla policy and 16.8\% versus 14.4\% for the action-space bridge with two chunks, and the same failure modes (capture for A$\to$B, freezing for B$\to$A).

\subsection*{Statistical analysis}
Success, capture, and freezing are proportions of trials and are reported with their trial numbers; intervals of success rates are 95\% Wilson score intervals. Because the flow noise is seeded by the initial state and the query index, methods are compared on paired counterfactual trials from the same initial state, switch step, and noise, and paired differences in success are tested with the two-sided exact McNemar test on the discordant pairs; intervals of paired differences in success rate are Newcombe intervals. For the self-trained policies, the comparisons were fixed before their test runs (pre-registered): pairs are formed within each training run and pooled over the runs. Bands and intervals of the transmission ratio and of the attention mass are percentile bootstrap intervals over rollouts (2,000 resamples per time bin). The language-free probe is a ridge-regularized linear classifier on standardized features, trained and tested on 20 random splits of the rollouts with one quarter held out at each time bin, and its band is a percentile bootstrap interval over the splits. All configurations were selected on the development split (initial states 40--49), and every test number comes from a single run.

\subsection*{Use of large language models}
A large language model (Claude, Anthropic) was used to polish the language of the manuscript and to write part of the code. The authors checked all text and code produced with its assistance and take full responsibility for the manuscript.

\section*{Data availability}
The LIBERO benchmark\cite{libero} and the released $\pi_0$ and $\pi_{0.5}$ checkpoints\cite{pi0,pi05,openpi} used in this study are publicly available.

\section*{Code availability}
The code of \ours{}, the switching protocol, the attention wormhole, and the analysis scripts is publicly available on GitHub at \url{https://github.com/xtli12/WarmHole}.

\section*{Author contributions}
X.L. conceived the study, formulated the instruction black hole, developed \ours{}, implemented the code, performed the experiments, analyzed the results, prepared the figures and wrote the manuscript. C.L. and J.W. took part in the discussions of the method and the experiments and revised the manuscript. S.L. provided guidance on the research and revised the manuscript. M.L. supervised the project and revised the manuscript. All authors approved the final manuscript.

\section*{Competing interests}
The authors declare no competing interests.

\bibliography{reference}

\end{document}

\flushbottom
\maketitle
\thispagestyle{empty}

\section*{Supplementary Note 1: Proofs}
\noindent\textit{Proof of Proposition~1 (static imitation is blind to the black hole).}\quad
(i) For $\chi\in\Ccal$, $p_\Dcal(l\mid\chi)=\mathbf{1}[l=g(\chi)]$, so $H_\Dcal(L\mid\chi)=0$ and $I_\Dcal(A;L\mid\chi)\le H_\Dcal(L\mid\chi)=0$.
(ii) Since $\tilde\pi=\pi$ outside $\Ccal$,
$J_\Dcal(\tilde\pi)-J_\Dcal(\pi)=\E_\Dcal\big[\mathbf{1}\{\chi\in\Ccal\}\big(\ell(\tilde\pi(\cdot\mid\chi,L),A)-\ell(\pi(\cdot\mid\chi,L),A)\big)\big]$.
On $\Ccal$ we have $L=g(\chi)$ $\Dcal$-almost surely, hence $\tilde\pi(\cdot\mid\chi,L)=\pi(\cdot\mid\chi,g(\chi))=\pi(\cdot\mid\chi,L)$ and the integrand vanishes almost surely. For the flow-matching loss, $\ell$ depends on $(\chi,l)$ only through the velocity field $v(\cdot,\cdot\mid\chi,l)$, and the same argument applies with the deafened field $\tilde v(\cdot,\cdot\mid\chi,l)=v(\cdot,\cdot\mid\chi,g(\chi))$.
(iii) For $\chi\in\Ccal$, $\tilde\pi(\cdot\mid\chi,l)$ does not depend on $l$, so $\Delta_{\tilde\pi}(\chi)=\E_{\tilde\pi}[z_1\mid\chi,\lnew]-\E_{\tilde\pi}[z_1\mid\chi,\lold]=0$, and $\rho_{\tilde\pi}(\chi)=0$ whenever $d^\star(\chi)\neq0$. \hfill$\square$

\noindent\textit{Proof of Proposition~2 (relocation).}\quad
(i) Conditioning the objective on the state, $J_\Dcal(\pi)=\E_\chi\,\E_{(L,A)\mid\chi}\big[\ell(\pi(\cdot\mid\chi,L),A)\big]$. For $\chi\in\Ccal$, $L=g(\chi)$ almost surely, so the inner expectation involves $\pi(\cdot\mid\chi,l)$ only at $l=g(\chi)$ and equals the expected loss of $\pi(\cdot\mid\chi,g(\chi))$ against the expert's action distribution $\pi^{\rm E}(\cdot\mid\chi,g(\chi))$. For the negative log-likelihood, this is minimized exactly by $\pi(\cdot\mid\chi,g(\chi))=\pi^{\rm E}(\cdot\mid\chi,g(\chi))$ (Gibbs' inequality); for conditional flow matching, by the conditional expectation of the target velocity, whose flow transports the noise to $\pi^{\rm E}(\cdot\mid\chi,g(\chi))$. Since the objective decomposes over states, a minimizer of $J_\Dcal$ attains these minima for $\Dcal$-almost every $\chi\in\Ccal$.
(ii) By definition of the deafened policy, $\tilde\pi(\cdot\mid\chi,l)=\pi(\cdot\mid\chi,g(\chi))$ for every $l$ and $\chi\in\Ccal$, which equals $\pi^{\rm E}(\cdot\mid\chi,g(\chi))$ by (i), and $J_\Dcal(\tilde\pi)=J_\Dcal(\pi)$ by Proposition~1(ii). The consequence follows by taking $\chi=y^*$ with $g(y^*)=\lnew$: with instruction $\lnew$, every minimizer acts as the expert of $\lnew$ by (i), and a deaf minimizer does so for any instruction by (ii). \hfill$\square$

\section*{Supplementary Note 2: Additional results on the LIBERO-Spatial pair}
\begin{figure}[!htp]
\centering
\includegraphics[width=\textwidth]{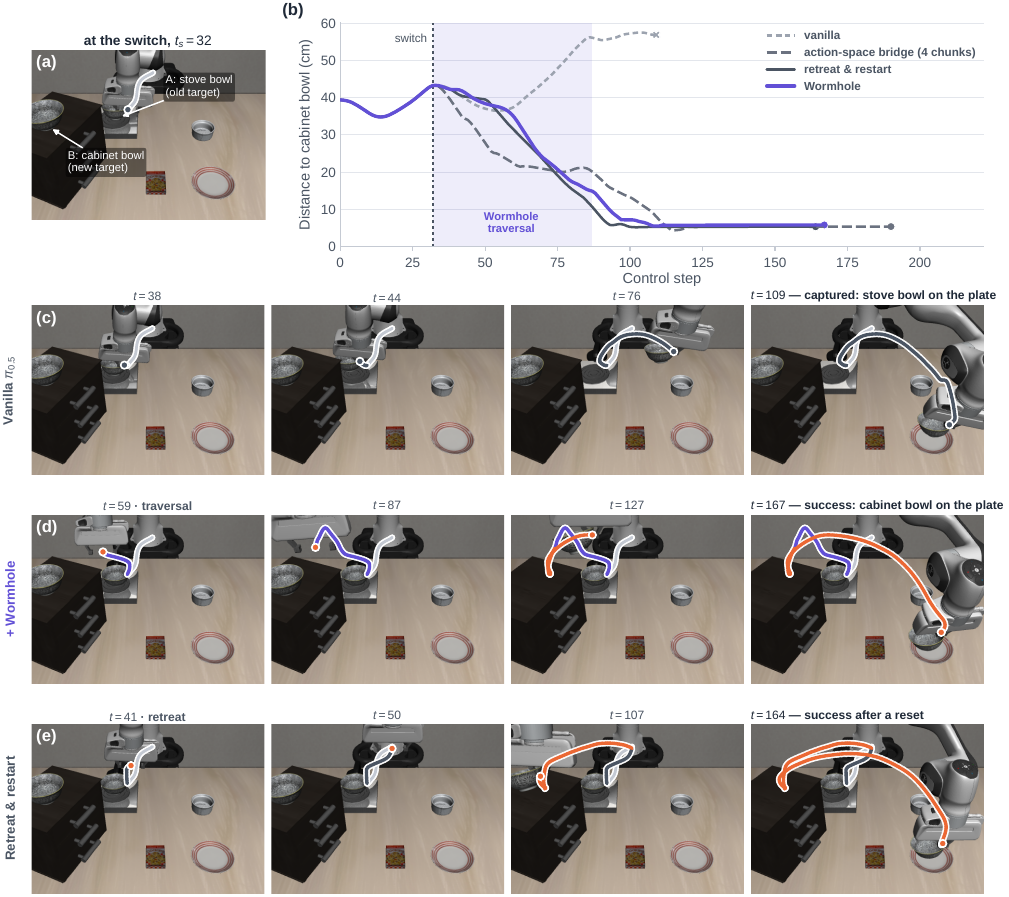}
\caption{\textbf{One switch, three responses.} A paired example on the LIBERO-Spatial pair with $\pi_{0.5}$, rendered from the simulator with the end-effector path projected into the agent-view camera. All rollouts start from initial state 0 with the same seeded noise and coincide until the instruction is switched from A (fetch the bowl on the stove) to B (fetch the bowl on the wooden cabinet) at $t_s=32$, when the gripper is closing on the stove bowl (light gray path). (a) The shared state at the switch; the two bowls are identical, so only language separates A from B. (b) Distance from the end effector to the cabinet bowl; the shaded band marks the traversal of \ours{}. (c) The vanilla policy ignores the switch and puts the stove bowl on the plate (capture). (d) \ours{} opens the gripper, lifts, transits, and settles at the far mouth in the approach phase of B (violet, 55 steps); the unmodified policy, queried under B, then grasps the cabinet bowl and places it on the plate (orange). (e) Retreat-and-restart opens the gripper and returns to the initial pose with a scripted controller (dark gray), then restarts under B. The action-space bridge with four chunks (dashed in (b)) also succeeds in this rollout, after a longer detour.}
\label{fig:qualitative}
\end{figure}

\noindent\textbf{One switch, three responses.} Supplementary Fig.~\ref{fig:qualitative} shows one paired rollout of $\pi_{0.5}$ from initial state 0 under the vanilla policy, \ours{}, and retreat-and-restart; the rollout of Fig.~1(a) of the main text is the same one.

\begin{figure}[!htp]
\centering
\includegraphics[width=\textwidth]{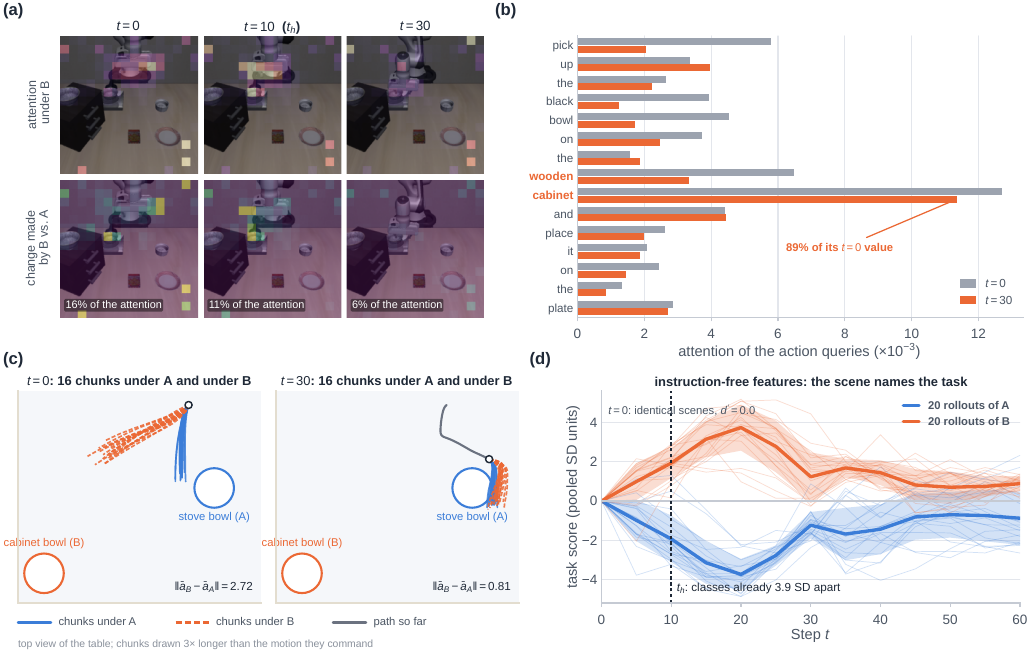}
\caption{\textbf{Inside the policy along one rollout} ($\pi_{0.5}$, spatial pair, no-switch rollout of A from initial state 0; at every query step, the recorded observation is re-run under A and under B with four shared noise samples). (a) Attention of the action queries over the 256 patches of the third-person image under B (top; mean over layers, heads, integration steps, and noise samples) and its change with respect to A, $|\alpha(B)-\alpha(A)|$ (bottom), at $t=0$, 10, and 30; the label gives the summed change as a share of the attention mass on this image (19\%, 7\%, and 13\% at $t=5$, 20, and 40). (b) Attention on the words of B at $t=0$ and $t=30$ (the beginning-of-sequence token, which absorbs about 0.3 of the attention at every step, is omitted). (c) Top view of the table with the end-effector motion commanded by 16 chunks sampled under A and under B at $t=0$ and $t=30$, integrated from the current position with the calibrated scale of 1.21\,cm per action unit and step (median endpoint error 12\%) and drawn three times longer for legibility; $\|\bar a_B-\bar a_A\|$ is the distance between the mean chunks in action units. (d) Projection of the instruction-free backbone features of 20 rollouts of A and 20 of B onto the difference of the class means at each step, in units of the pooled standard deviation (thin lines: rollouts; thick lines and bands: class means $\pm$ one standard deviation). The scenes are identical at $t=0$, and at $t_h$ the classes are already 3.9 standard deviations apart.}
\label{fig:inside}
\end{figure}

\noindent\textbf{Inside one rollout.} Supplementary Fig.~\ref{fig:inside} opens the policy along a single no-switch rollout of $\pi_{0.5}$ under $\lold$ (initial state 0), the rollout from which the switch of Supplementary Fig.~\ref{fig:qualitative} is issued. At each query step, we run the policy on the recorded observation under $\lold$ and under $\lnew$ with shared noise and record the attention of the action queries over the prefix tokens. At $t=0$, $\lnew$ changes where the action queries look in the third-person image---the summed absolute change amounts to 16\% of the attention mass on that image---and it changes what they command: the 16 sampled chunks head for the stove bowl under $\lold$ and for the cabinet bowl under $\lnew$ (Supplementary Fig.~\ref{fig:inside}(c); the mean chunks differ by 2.72 in action units). At $t=30$, two steps before the switch of Supplementary Fig.~\ref{fig:qualitative}, the instruction is still read---``cabinet'' remains the most attended word of $\lnew$, at 89\% of its attention at $t=0$---but it barely changes where the policy looks (6\% of the image attention mass) and what it commands: both sets of chunks enter the stove bowl, and their mean difference has shrunk to 0.81 ($\pi_0$, along its own rollout: from 2.84 to 0.81). In the same window, the language-free features of 20 rollouts per instruction fork at $t_h$ (Supplementary Fig.~\ref{fig:inside}(d)): the scene has begun to name the task by itself. This is the black hole at the level of a single state: the words of $\lnew$ still reach the action expert, but the action is already determined by the state.

\noindent\textbf{What \ours{} computes at the switch.} Supplementary Fig.~\ref{fig:runtime} traces Algorithm~1 of the main text through the rollout of Supplementary Fig.~\ref{fig:qualitative}. At the switch, the gripper is 25\,cm from the approach phase of B. The approach-phase entries of B trade the traversal needed to reach them against the steps that remain from them (Supplementary Fig.~\ref{fig:runtime}(b)). The far mouth chosen online is essentially the $b$-weighted entry: the far mouth recomputed offline from the recorded state without any exits lies within 0.1\,mm of it, i.e., the plan is close to the product of its marginals, as expected when the exits form a single mode deep in the black hole (Methods of the main text). It lies 29 steps into an execution of B, with 84 steps to go. The traversal reaches it after 55 steps, and from there the unmodified policy, queried under B, grasps the cabinet bowl and places it on the plate (Supplementary Fig.~\ref{fig:runtime}(a)). Supplementary Fig.~\ref{fig:runtime}(c) shows where each method leaves the robot: the retreat and \ours{} bring the gripper within 2\,cm of the approach phase of B before the policy acts on its own, whereas the action-space bridge hands over 16\,cm away and the vanilla policy comes within 3\,cm only in passing, while it carries the stove bowl to the plate.

\begin{figure}[!htp]
\centering
\includegraphics[width=\textwidth]{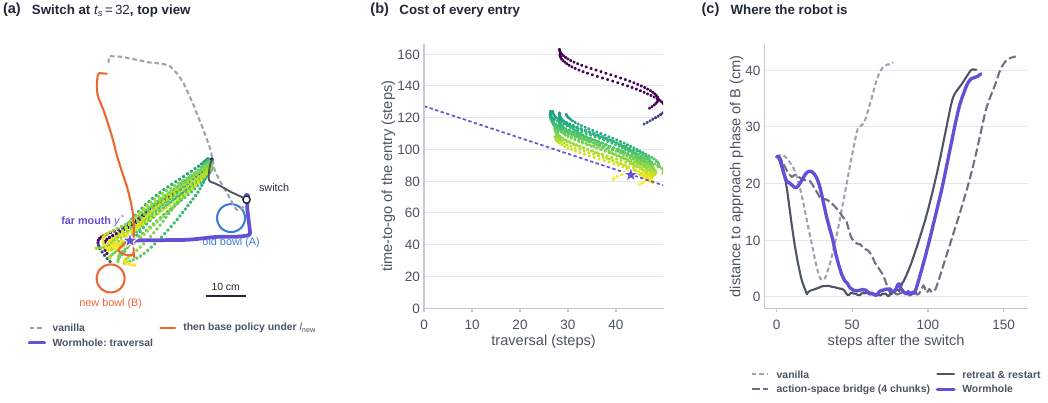}
\caption{\textbf{What \ours{} computes at a committed switch} ($\pi_{0.5}$, spatial pair A$\to$B, initial state 0, $t_s=32$: the rollout of Supplementary Fig.~\ref{fig:qualitative}). (a) Top view: the approach-phase entries of B (dots, colored by the expected number of steps to complete B through them, traversal plus time-to-go; yellow is cheapest), the far mouth $y^*$ (star), the executed traversal of \ours{} (violet) followed by the base policy under B (orange), and the vanilla rollout from the same state (gray). (b) The trade-off behind the far mouth: traversal steps versus time-to-go of every entry, with the iso-cost line through the selected mouth. (c) Distance of the end effector to the approach phase of B after the switch, for the four methods from the same state; it rises again once B is grasped and carried to the plate.}
\label{fig:runtime}
\end{figure}

\begin{figure}[!htp]
\centering
\includegraphics[width=\textwidth]{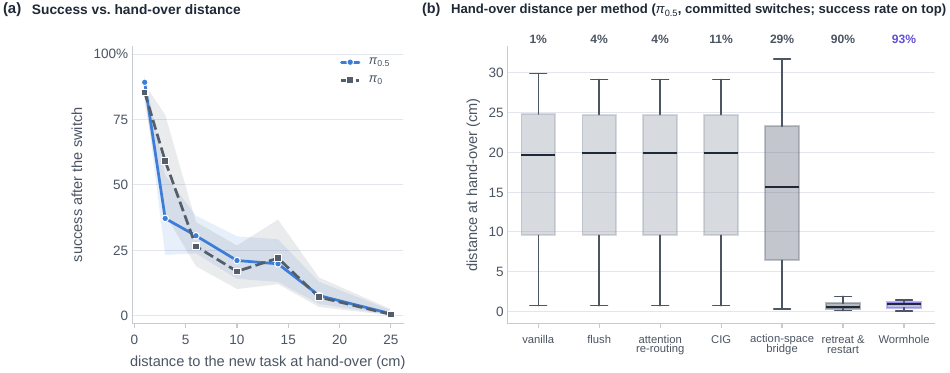}
\caption{\textbf{Relocation decides the escape} (committed switches on the LIBERO-Spatial pair, all methods). (a) Success after the switch as a function of the distance between the end effector at hand-over and the approach phase of successful executions of the new instruction, pooled over all methods (1360 trials for $\pi_{0.5}$ and 960 for $\pi_0$; 95\% Wilson bands). (b) Hand-over distance per method for $\pi_{0.5}$ (boxes: interquartile range; whiskers: 1.5 interquartile ranges; the action-space bridge pools two, four, and six chunks), with the committed success rate on top.}
\label{fig:relocation}
\end{figure}

\noindent\textbf{Relocation decides the escape.} Supplementary Fig.~\ref{fig:relocation} tests prediction (P3) across every method and every committed switch of the spatial pair, 1360 trials for $\pi_{0.5}$ and 960 for $\pi_0$. We measure the distance from the end effector at hand-over to the approach phase of successful executions of $\lnew$; hand-over is the end of the traversal or the retreat, and the switch step for methods that do not move the robot themselves. Success falls steeply with this distance, almost identically for both backbones: hand-overs within 3\,cm succeed in 87.7\% ($\pi_{0.5}$) and 84.3\% ($\pi_0$) of the trials, those beyond 15\,cm in 3.4\% and 2.5\% (Supplementary Fig.~\ref{fig:relocation}(a)). The methods separate along the same axis (Supplementary Fig.~\ref{fig:relocation}(b)): the inference-time remedies leave the robot where the vanilla policy is, a median of 20\,cm from the new task, the action-space bridge brings it to 9--18\,cm, and only the two relocations hand over within 3\,cm, \ours{} in 98.1\% ($\pi_{0.5}$) and 98.8\% ($\pi_0$) of its trials and the retreat in 99.4\% and 100\%.

\section*{Supplementary Note 3: Training-free remedies and ablation study}
Supplementary Table~\ref{tab:remedies} lists the training-free remedies of the main text and Supplementary Table~\ref{tab:ablation} the ablation study, both on $\pi_{0.5}$ with initial states 0--9 (20 early and 80 committed trials per row). For reference, both tables also list the methods that move the robot on the same trials: the action-space bridge, \emph{retreat-and-restart}, which opens the gripper, lifts, and returns the arm to its initial pose with a scripted controller before the policy executes $\lnew$, and \ours{}. The retreat moves the state back across the horizon, to a configuration from which the task cannot be read without language (Fig.~4(a) of the main text), so the instruction is heard again---the converse of Proposition~1; \ours{} moves the state forward into the committed region of the new task, where, by Proposition~2, the state itself selects the new task.

\begin{table}[!htp]
\footnotesize
\centering
\caption{\textbf{Training-free remedies on $\pi_{0.5}$} (spatial pair, initial states 0--9; 20 early and 80 committed trials per row). Columns as in Table~1 of the main text, without late switches. The rows below the line move the robot and are listed for reference on the same trials.}
\label{tab:remedies}
\renewcommand{\arraystretch}{1.25}
\setlength{\tabcolsep}{3pt}
\begin{tabularx}{\textwidth}{@{}l Y{0.84}Y{1.76}Y{0.84}Y{0.84}Y{1.2}Y{0.84}Y{0.68}@{}}
\toprule
\headrow
Method & Early\,\up & Committed\,\up & Capture\,\down & Freeze\,\down & Completion\,\down & Disc.\,\down & Jerk\,\down \\
\midrule
Replace (vanilla) & 95.0 & 1.2 {\scriptsize[0.2, 6.7]} & 66.2 & 32.5 & 133$^\dagger$ & \best{0.09} & 0.12 \\
Flush & \best{100.0} & 3.8 {\scriptsize[1.3, 10.5]} & 71.2 & 25.0 & 158$^\dagger$ & 0.34 & 0.10 \\
Attention re-routing ($\beta=2$) & 95.0 & 3.8 {\scriptsize[1.3, 10.5]} & 68.8 & 27.5 & 122$^\dagger$ & 0.35 & 0.10 \\
Attention re-routing ($\beta=5$) & 95.0 & 5.0 {\scriptsize[2.0, 12.2]} & 47.5 & 47.5 & 144$^\dagger$ & 0.39 & 0.09 \\
CIG ($w=1$) & 85.0 & 11.2 {\scriptsize[6.0, 20.0]} & 32.5 & 56.2 & 153$^\dagger$ & 0.56 & 0.15 \\
CIG ($w=3$) & 70.0 & 11.2 {\scriptsize[6.0, 20.0]} & 18.8 & 70.0 & 129$^\dagger$ & 0.76 & 0.24 \\
\midrule
Action-space bridge (4 chunks) & 90.0 & 32.5 {\scriptsize[23.2, 43.4]} & 30.0 & 37.5 & 146 & 0.33 & 0.16 \\
Retreat-and-restart (scripted reset) & \best{100.0} & 88.8 {\scriptsize[80.0, 94.0]} & \best{0.0} & 11.2 & \best{137} & 1.51 & 0.30 \\
\oursrow \ours{} & 95.0 & \best{95.0 {\scriptsize[87.8, 98.0]}} & \best{0.0} & \best{5.0} & 142 & 0.32 & \best{0.09} \\
\bottomrule
\end{tabularx}
\end{table}

\begin{table}[!htp]
\footnotesize
\centering
\caption{\textbf{Ablation study on $\pi_{0.5}$} (spatial pair, initial states 0--9; 20 early and 80 committed trials per row). Columns as in Supplementary Table~\ref{tab:remedies}. The lower block lists the action-space bridge of our earlier design on the same trials, without bold.}
\label{tab:ablation}
\renewcommand{\arraystretch}{1.2}
\setlength{\tabcolsep}{2.5pt}
\begin{tabularx}{\textwidth}{@{}l l Y{0.74}Y{1.7}Y{0.9}Y{0.84}Y{1.26}Y{0.8}Y{0.76}@{}}
\toprule
\headrow
Component & Variant & Early\,\up & Committed\,\up & Capture\,\down & Freeze\,\down & Completion\,\down & Disc.\,\down & Jerk\,\down \\
\midrule
Mouth B & entries from all phases (incl. after the grasp) & 95.0 & 1.2 {\scriptsize[0.2, 6.7]} & \best{0.0} & 98.8 & 115$^\dagger$ & \best{0.32} & 0.08 \\
 & phase margin $m=20$ & 95.0 & 96.2 {\scriptsize[89.5, 98.7]} & \best{0.0} & 3.8 & 136 & \best{0.32} & 0.09 \\
Throat & nearest entry ($\lambda=0$, no time-to-go) & 95.0 & \best{97.5 {\scriptsize[91.3, 99.3]}} & \best{0.0} & \best{2.5} & 141 & \best{0.32} & 0.10 \\
 & no exits, no OT ($b$-weighted entry) & 95.0 & 92.5 {\scriptsize[84.6, 96.5]} & \best{0.0} & 7.5 & 142 & \best{0.32} & 0.09 \\
Traversal & no lift, no clearance & 95.0 & \best{97.5 {\scriptsize[91.3, 99.3]}} & \best{0.0} & \best{2.5} & \best{122} & \best{0.32} & \best{0.07} \\
 & no rate limit & 95.0 & 93.8 {\scriptsize[86.2, 97.3]} & \best{0.0} & 6.2 & 132 & 1.50 & 0.21 \\
\oursrow Full & \ours{} & 95.0 & 95.0 {\scriptsize[87.8, 98.0]} & \best{0.0} & 5.0 & 142 & \best{0.32} & 0.09 \\
\midrule
Action-space & 2 chunks (closed form) & 100.0 & 28.7 {\scriptsize[20.0, 39.5]} & 32.5 & 38.8 & 147 & 0.33 & 0.16 \\
bridge & 4 chunks & 90.0 & 32.5 {\scriptsize[23.2, 43.4]} & 30.0 & 37.5 & 146 & 0.33 & 0.16 \\
 & 6 chunks & 90.0 & 26.2 {\scriptsize[17.9, 36.8]} & 22.5 & 51.2 & 179 & 0.32 & 0.16 \\
 & learned bridge head (2 chunks) & 100.0 & 25.0 {\scriptsize[16.8, 35.5]} & 31.2 & 43.8 & 137 & 0.73 & 0.11 \\
\bottomrule
\end{tabularx}
\end{table}

\noindent\textbf{Further variants of the action-space bridge} on the same trials do no better than the rows of Supplementary Table~\ref{tab:ablation}: 23.8\% without OT, 26.2\% without anchoring, 31.2\% without its three-step ramp (switch discontinuity 0.86), 16.2\% with one chunk, and 6.2\% with entries drawn without state compatibility. The first training run of the learned bridge head produced non-finite weights because the plain-domain Sinkhorn iterations underflowed for training states far from every entry; the 140 episodes run with those heads are discarded. The head reported in Supplementary Table~\ref{tab:ablation} was retrained with log-domain Sinkhorn iterations and evaluated in a separate run of the same 140 episodes. Over the 315 committed traversals of the test split and the 184 traversals on LIBERO-Goal, the far mouth chosen with OT lies within 0.5\,mm of the $b$-weighted entry (median 0.01\,mm).

\section*{Supplementary Note 4: Self-trained attention wormhole, diagnostics}
\begin{figure}[!htp]
\centering
\includegraphics[width=\textwidth]{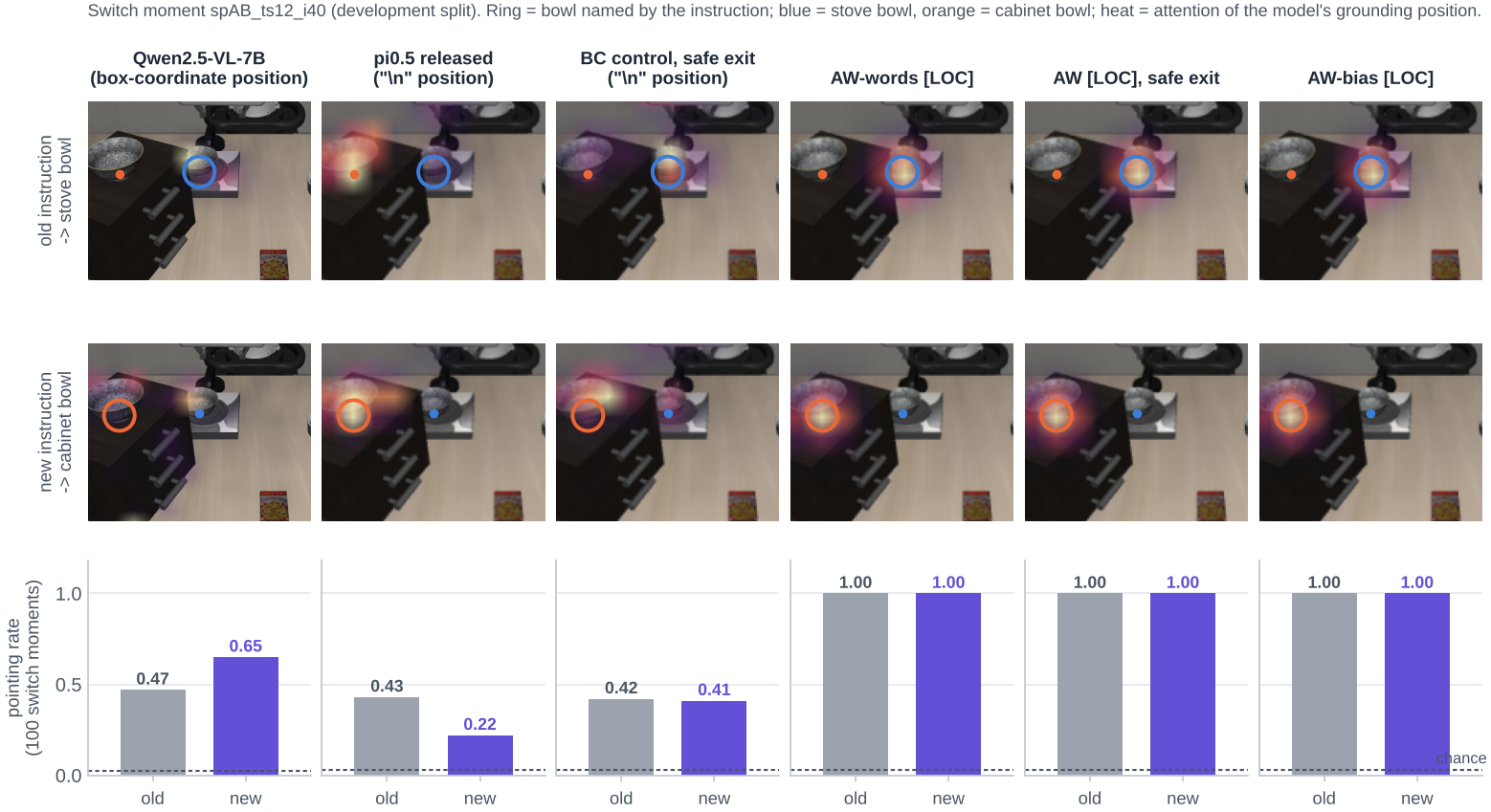}
\caption{\textbf{Where the models look when the instruction changes} (development split, the switch moment $t_s=12$ of initial state 40, A$\to$B). Rows: attention under the old and the new instruction; ring: the bowl named by the instruction. Columns: Qwen2.5-VL-7B at the position that writes the box coordinates (mean of its three best heads), the released $\pi_{0.5}$ and the BC policy trained on the safe-exit data at the instruction position (three best heads), and the [LOC] token of three self-trained variants (AW-words, AW trained on the safe-exit data, AW-bias; heads 0--3 of layers 6--13). Bars: rate at which the arg-max patch lies within 1.5 patches of the named bowl over 100 switch moments (chance 0.028).}
\label{fig:distill}
\end{figure}

\noindent\textbf{Where the models look.} Supplementary Fig.~\ref{fig:distill} compares, at the switch moments of the development split, where Qwen2.5-VL-7B, the released $\pi_{0.5}$, the BC policy, and the self-trained variants of the main text look when the instruction changes. The [LOC] token points to the referred bowl at all 100 switch moments before and after the switch, in every AW variant including the two trained on the safe-exit data, whereas the best heads of the BC policies point to it in 0.42 and 0.31 of them (original data) and 0.42 and 0.41 (safe-exit data).

\noindent\textbf{Causal diagnostics on the development split} (AW without augmentation unless stated). Hiding [LOC] from the action expert at the switch drops committed success from 95.0\% to 0.0\% in both directions, mostly by freezing, whereas removing only the attention bias leaves 95.0\%: the key and value of [LOC] carry the instruction. The bias alone is a weaker carrier: AW-bias follows 81.2\% of committed and 61.2\% of late test switches, and on the development split it loses every B$\to$A switch once its bias is removed. When the action expert may also read the instruction words (AW-words), it ignores the wormhole: it still succeeds in 93.8\% of committed switches with [LOC] and the bias switched off, and pairing the old words with the new [LOC] sends every trial to the old bowl. The same diagnostics on the two AW policies trained on the safe-exit data give the same picture: hiding [LOC] from the action expert drops committed success on the development split from 96.2\% to 20.0\% (run 0) and from 91.2\% to 5.0\% (run 1), to 0.0\% in every B$\to$A trial of both runs, while 38--44\% of the trials are captured by the old bowl; the remaining A$\to$B successes (40\% and 10\%) are a default toward the cabinet bowl that happens to be the new target, not a correction. Removing only the attention bias leaves the success counts unchanged (77 and 73 of 80 trials). The variation between training runs is larger for AW than for BC on the original data (212 versus 196 successes of the two AW runs on the paired committed and late trials, $p=0.017$; 196 versus 195 for BC), and the first run alone showed a significant difference over BC (212 versus 196, $p=0.026$) that the second did not reproduce (196 versus 195, $p=1$). With the safe-exit data, run 0 of AW reaches 94.4\% of committed and 81.2\% of late test switches.

\noindent\textbf{Attempts that did not help.} Five further attempts to improve the trained policy were evaluated on the development split under rules fixed before each run and were not tested, because they lowered or did not raise development success: self-imitation from switch self-play (AW retrained on its own successful switches: 589 versus 629 of 720 development trials), contrastive instruction guidance at inference ($w=0.5$ and $1.0$: 433 and 344 for AW, 527 and 484 for BC), and three relabeling schemes after the switch (on-policy counterfactual correction on the attractor states and a corridor teacher that first lifts the arm, both of which yielded too few successful teacher episodes to train on, and relabeling the student's own committed states at the switch, which gave 592). The safe-exit teacher hands back at a median elbow margin of 1.10\,rad, against 0.86\,rad for the original one.

